\documentclass{article}

\usepackage[final]{neurips_2024}

\usepackage[utf8]{inputenc} 
\usepackage[T1]{fontenc}    
\usepackage{hyperref}       
\usepackage{url}            
\usepackage{booktabs}       
\usepackage{amsfonts}       
\usepackage{nicefrac}       
\usepackage{microtype}      
\usepackage{xcolor}         

\usepackage{graphicx}
\usepackage{subfigure}
\usepackage{booktabs} 

\usepackage{amsmath}
\usepackage{amssymb}
\usepackage{mathtools}
\usepackage{amsthm}

\usepackage{pdflscape}
\usepackage{enumitem}
\usepackage{rotating} 

\usepackage{amsmath,amsfonts,bm}

\def\eqref#1{equation~\ref{#1}}

\def\1{\bm{1}}

\def\vv{{\bm{v}}}

\def\vx{{\bm{x}}}

\def\mC{{\bm{C}}}

\def\mE{{\bm{E}}}

\def\mH{{\bm{H}}}

\def\mM{{\bm{M}}}
\def\mN{{\bm{N}}}

\def\mP{{\bm{P}}}

\def\mV{{\bm{V}}}
\def\mW{{\bm{W}}}
\def\mX{{\bm{X}}}

\DeclareMathAlphabet{\mathsfit}{\encodingdefault}{\sfdefault}{m}{sl}
\SetMathAlphabet{\mathsfit}{bold}{\encodingdefault}{\sfdefault}{bx}{n}

\usepackage{wrapfig}
\usepackage{url}
\definecolor{color1}{HTML}{006EB8}
\definecolor{color2}{HTML}{009B55}
\definecolor{color3}{HTML}{00A99A}
\definecolor{color4}{HTML}{3C8031}
\definecolor{color5}{HTML}{006795}
\definecolor{color6}{HTML}{00AEB3}
\definecolor{mygray}{gray}{0.9}
\usepackage{hyperref}
\hypersetup{
  colorlinks   = true, 
  urlcolor     = magenta, 
  linkcolor    = magenta, 
  citecolor   = color5 
}

\usepackage{booktabs}
\usepackage{graphicx}
\usepackage{xcolor}
\usepackage{subcaption}
\usepackage{tabularx}
\usepackage[export]{adjustbox}
\usepackage{xspace}
\usepackage{makecell}
\usepackage{multirow}
\usepackage{booktabs}
\usepackage[capitalize,noabbrev]{cleveref}
\crefformat{section}{\S#2#1#3}
\crefformat{subsection}{\S#2#1#3}
\crefformat{subsubsection}{\S#2#1#3}

\newcommand{\model}{LLaMA\xspace}

\newcommand{\appendixref}[1]{\hyperref[#1]{Appendix~\ref*{#1}}}

\theoremstyle{plain}

\theoremstyle{definition}

\theoremstyle{remark}

\usepackage[textsize=tiny]{todonotes}
\title{
HuRef: HUman-REadable Fingerprint\\ for Large Language Models}

\author{Boyi Zeng$^1$, Lizheng Wang$^2$, Yuncong Hu$^2$, Yi Xu$^2$\\ \textbf{Chenghu Zhou$^3$,} \textbf{Xinbing Wang$^2$,} \textbf{Yu Yu$^2$,} \textbf{Zhouhan Lin}$^{1}$\thanks{Zhouhan Lin is the corresponding author.}\\
$^1$LUMIA Lab, Shanghai Jiao Tong University  \\$^2$Shanghai Jiao Tong University, $^3$Chinese Academy of Sciences\\
\texttt{boyizeng@sjtu.edu.cn
      $^*$lin.zhouhan@gmail.com} } 

\begin{document}

\maketitle
\setcounter{footnote}{0}
\begin{abstract}
Protecting the copyright of large language models (LLMs) has become crucial due to their resource-intensive training and accompanying carefully designed licenses. However, identifying the original base model of an LLM is challenging due to potential parameter alterations. In this
study, we introduce HuRef, a human-readable fingerprint for LLMs that uniquely identifies the base model without interfering with training or exposing model parameters to the public.
We first observe that the vector direction of LLM parameters remains stable after the model has converged during pretraining, 
with negligible perturbations through subsequent training steps, including continued pretraining, supervised fine-tuning, and RLHF, 
which makes it a sufficient condition
to identify the base model.
The necessity is validated by continuing to train an LLM with an extra term to drive away the model parameters' direction and the model becomes damaged. However, this direction is vulnerable to simple attacks like dimension permutation or matrix rotation, which significantly change it without affecting performance. To address this, leveraging the Transformer structure, we systematically analyze potential attacks and define three invariant terms that identify an LLM's base model. 
Due to the potential risk of information leakage, we cannot publish invariant terms directly. Instead, we map them to a Gaussian vector using an encoder, then convert it into a natural image using StyleGAN2, and finally publish the image. In our black-box setting, all fingerprinting steps are internally conducted by the LLMs owners. To ensure the published fingerprints are honestly generated, we introduced Zero-Knowledge Proof (ZKP).
Experimental results across various LLMs demonstrate the effectiveness of our method.\footnote{The code is available at \url{https://github.com/LUMIA-Group/HuRef}.}  
\end{abstract}

\section{Introduction}
Large language models (LLMs) have become the foundation models in many scenarios of artificial intelligence. As training an LLM from scratch consumes a huge amount of computation and data resources and the trained LLM needs to be carefully protected from malicious use, the parameters of the LLMs become a crucial property to protect, for both commercial and ethical reasons. As a result, many of the LLMs are open-sourced with carefully designed licenses to reject commercial use~\citep{touvron2023llama,taylor2022galactica} or requiring an apply-and-approval process~\citep{touvron2023llama2,zhang2022opt,falcon,baichuan,2023internlm,zheng2023codegeex}, let alone some LLMs are not open-sourced entirely~\citep{openaichatgpt,OpenAI2023GPT4TR,brown2020language,wu2023bloomberggpt,chowdhery2022palm,hoffmann2022training}.

At the core of protecting LLMs from unauthorized use is to identify the base model of a given LLM. However, different from other forms of property such as software or images, protecting LLMs is a novel problem with unique challenges. First, the base model usually needs to be fine-tuned or even continued pretraining to be applied to downstream tasks, resulting in parameter updates that make the resulting model different from the original base model, which makes it disputable to identify the base model. Second, many of the popular LLMs are not releasing their parameters, leaving the identification in a black-box setting. Third, different from previous smaller-scale neural networks that are only trained for specific tasks, LLMs are usually targeted for enormous forms of tasks that are not yet defined during pretraining. This has made the watermarking methods for traditional neural networks~\citep{adi2018turning,xiang2021protecting,yadollahi2021robust} not suited in this case, especially under extensive subsequent training steps.
\begin{figure*}[t]
\includegraphics[width=\textwidth]{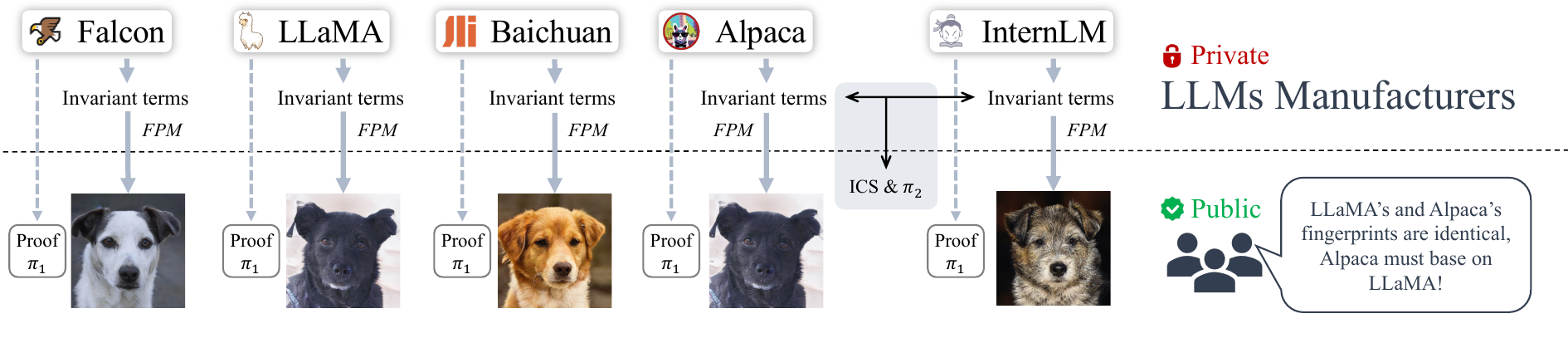}
\captionsetup{font=small}
\caption{
An illustrative framework for LLM protection with fingerprints. The LLM manufacturers compute invariant terms internally and feed them into the fingerprinting model (FPM\protect\footnotemark) to generate a fingerprint image. This image is then released to the public along with zero-knowledge proofs ($\pi_1$), allowing for intuitive identification of shared base models through the fingerprint images. We also provide a limited one-to-one quantitative comparison scheme (ICS \& $\pi_2$) as a complement.
Zero-Knowledge Proof guarantees the reliability of the fingerprints and comparison results, without interfering with LLM training or revealing model parameters to the public.
} 
\vspace{-4mm}
\label{fig:overall_logic}
\end{figure*}
\footnotetext{FPM is open-sourced, as all LLM manufacturers need to use the same one. We have placed it on the private side in~\autoref{fig:overall_logic} solely because the fingerprinting process is private.}

In this work, we propose a novel way to overcome the aforementioned challenges by proposing a method that reads part of the model parameters and computes a fingerprint for each LLM without interfering with training or exposing model parameters to the public. The appearance of the fingerprint is closely dependent on the base model, and invariant to almost all subsequent training steps, including supervised fine-tuning (SFT), reinforcement learning with human feedback (RLHF), or even continue-pretraining with augmented vocabulary in a new language. 

The fingerprint is based on our observation that the vector direction of LLM parameters remains stable against various subsequent training steps after the model has converged during pretraining. This makes it a good indicator for base model identification. Empirically, the sufficiency of this correlation is elaborated in~\autoref{Sufficiency3.1}, while its necessity is presented in~\autoref{Necessity}.

Further, despite its stability towards training, the vector direction of the model parameter is vulnerable to some simple direct weight rearrangements that could significantly change the direction of parameter vectors without affecting the model's performance. We construct three invariant terms that are robust to these weight rearrangements by systematically analyzing possible rearrangements and leveraging the Transformer structure. This is elaborated in~\autoref{Invariant Terms}.


Moreover, we generate human-readable fingerprints by mapping the invariant terms into a Gaussian random vector through an encoder and then mapping the Gaussian vector to a natural image through an off-the-shelf image generation model, such as StyleGAN2~\citep{karras2020analyzing}. This generation offers a dual benefit of mitigating information leakage and making our fingerprints straightforward to decipher.  To ensure the published fingerprints are honestly generated, we also introduced Zero-Knowledge Proof (ZKP). This is elaborated in~\autoref{The Fingerprinting Model}.

With this fingerprinting approach, we can sketch an outline for protecting LLMs in~\autoref{fig:overall_logic}.

\section{Related Works}

There are two primary categories of related approaches.

\textbf{Post-hoc Detection} methods involve analyzing text generated by LLMs after its production. LLMDet~\citep{wu2023llmdet} calculates proxy perplexity by leveraging prior knowledge of the model's next-token probabilities. DetectGPT~\citep{mitchell2023detectgpt} uses model-predicted probabilities to identify passages generated by a specific LLM.~\cite{li2023origin} employs perplexity scores and intricate feature engineering. These methods are usually applicable to a specific LLM and could be affected by supervised fine-tuning (SFT) and continued pretraining. More recently,~\cite{sadasivan2023aigenerated} presented theoretical findings that for highly advanced AI human mimickers, even the best possible detection methods may only marginally outperform random guessing.

\textbf{Watermarking Techniques} can be divided into two main categories~\citep{boenisch2021systematic}. The first embeds watermarks or related information into the model parameters, such as explicit watermarking scheme~\citep{Uchida_2017} or leveraging another neural network~\citep{wang2020watermarking}, which could potentially affect model performance~\citep{wang2019attacks}. The second category focuses on inducing unusual prediction behavior in the model.~\cite{xiang2021protecting} explored embedding phrase triggers, and~\cite{gu2023watermarking} extended this approach to LLM; however, they are task-specific.~\cite{yadollahi2021robust} proposed a watermarking method but did not consider subsequent fine-tuning.~\cite{cryptoeprint:2023/763} proposed a cryptographic approach, but it is not robust to text editing.~\cite{kirchenbauer2023watermark} involved using pre-selected tokens
which inevitably alters the model prediction. These methods may turn out to be vulnerable to attacks on certain tokens, for example,~\cite{krishna2023paraphrasing} successfully evaded watermarking~\citep{kirchenbauer2023watermark}, GPTZero~\citep{Gptzero}, and OpenAI's text classifier~\citep{ai-text-classifier} using paraphrasing attacks.

Our work doesn't fall into any of the two categories since it is based on analyzing model weights post-hoc and relies on a wide spectrum of tokens. 

In~\autoref{Additional related works}, we provide a more extensive discussion of additional related works.
\begin{table*}[t]
\begin{minipage}{0.66\textwidth}  
\small
\centering
\setlength\tabcolsep{1pt}
\resizebox{\columnwidth}{!}{
\begin{tabular}{@{}l|cccccccccc|cccc@{}}
\toprule
\rotatebox{90}{\textbf{Model}} & \rotatebox{90}{\textcolor{blue}{Alpaca-Lora}} & \rotatebox{90}{\textcolor{blue}{Alpaca}} & \rotatebox{90}{\textcolor{blue}{Chinese-\model}} & \rotatebox{90}{\textcolor{blue}{Vicuna}} & \rotatebox{90}{\textcolor{blue}{Baize}} & \rotatebox{90}{\textcolor{blue}{MedAlpaca}} & \rotatebox{90}{\textcolor{blue}{Koala}}& \rotatebox{90}{\textcolor{blue}{WizardLM}} & \rotatebox{90}{\textcolor{blue}{MiniGPT-4}} & \rotatebox{90}{\textcolor{blue}{Chinese-Alpaca}} & \textcolor{red}{\rotatebox{90}{Baichuan}} & \textcolor{red}{\rotatebox{90}{OpenLLaMA}} & \textcolor{red}{\rotatebox{90}{InternLM}} & \textcolor{red}{\rotatebox{90}{LLaMA-2}} \\ \midrule
PCS & \textcolor{blue}{99.87} & \textcolor{blue}{99.91} & \textcolor{blue}{99.68} & \textcolor{blue}{99.80} & \textcolor{blue}{99.73} & \textcolor{blue}{99.90} & \textcolor{blue}{99.82}  & \textcolor{blue}{99.89} & \textcolor{blue}{99.70} & \textcolor{blue}{99.52} & \textcolor{red}{0.83} & \textcolor{red}{1.16} & \textcolor{red}{0.28} & \textcolor{red}{1.51} \\ \midrule
ICS & \textcolor{blue}{99.60} & \textcolor{blue}{99.95} & \textcolor{blue}{93.57} & \textcolor{blue}{99.42} & \textcolor{blue}{99.60} & \textcolor{blue}{99.86} & \textcolor{blue}{99.63} & \textcolor{blue}{99.89} & \textcolor{blue}{99.20} & \textcolor{blue}{91.35} & \textcolor{red}{0.32} & \textcolor{red}{0.32} & \textcolor{red}{0.03} & \textcolor{red}{3.16} \\ 
\bottomrule
\end{tabular}
}
\captionof{table}{The cosine similarities of model parameters (PCS) and invariant terms (ICS) between various LLMs w.r.t. the LLaMA-7B base model. All models are of the same size.}
\label{normal cosine similarities}
\end{minipage}
\begin{minipage}{0.34\textwidth}  
\centering
\includegraphics[width=\textwidth]{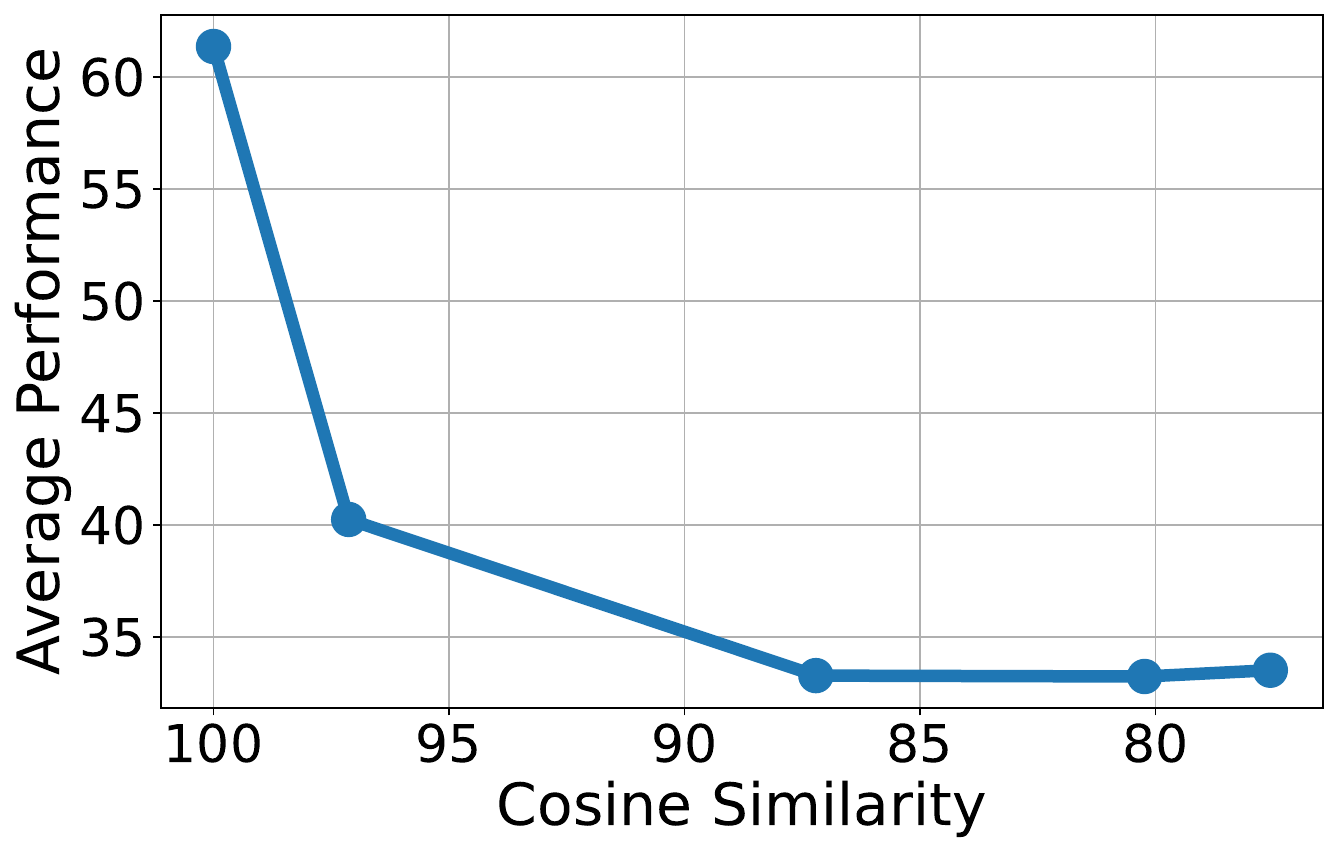}
\captionof{figure}{The model's performance quickly deteriorates as the PCS decreases.}
\label{fig:cosvsavg}
\end{minipage}
\end{table*}

\section{Vector Direction of LLM Parameters and the Invariant Terms}\label{Vector Direction and Invariant Terms}

\subsection{Using Vector Direction of LLM Parameters to Identify the Base Model}\label{normal cs}

We can flatten all weight matrices and biases of an LLM into vectors and concatenate them together into a single huge vector.
In this subsection, we are going to show how the direction of this vector could be used to determine the base model by empirically showing its sufficiency and necessity.

\subsubsection{Sufficiency}\label{Sufficiency3.1}

For sufficiency, we compute the cosine similarities between a base model LLaMA-7B and various of its offspring models, as well as other independently pretrained LLMs~\citep{openlm2023openllama} that are of the same size.~\autoref{normal cosine similarities} shows a wide spectrum of models that inherit the LLaMA-7B base model, whose subsequent training processes involve various training paradigms, such as SFT~\citep{alpaca,xu2023baize,zheng2023judging,geng2023easylm,xu2023wizardlm,han2023medalpaca}, SFT with LoRA~\citep{alpaca-lora} and extensive continued pretraining in a new language~\citep{chinese-llama-alpaca}, 
extending to new modalities~\citep{zhu2023minigpt}, etc. We detail the subsequent training settings of these models in Appendix~\autoref{tab:all offspring model descriptions}.

Regardless of their various subsequent training setting, we can figure that all of these models show almost full scores in cosine similarity, largely preserving the base model's parameter vector direction. On the other hand, the models that are trained independently appear to be completely different in parameter vector direction, showing almost zero cosine similarity with the LLaMA-7B model. 

These observations indicate that a high cosine similarity between the two models highly suggests that they share the same base model, and vice versa. 

\subsubsection{Necessity}\label{Necessity}
From the necessity perspective, we want to verify if the base model's ability can still be preserved when the cosine similarity is intentionally suppressed in subsequent training steps. To this end, we inherit the LLaMA-7B base model and interfere with the Alpaca's SFT process by augmenting the original SFT loss with an extra term that minimizes the absolute value of cosine similarity. i.e. $L_A = \frac{\left|\langle \displaystyle \mV_A ,\displaystyle \mV_{base} \rangle \right| }{|\displaystyle \mV_A| |\displaystyle \mV_{base}|}$. Here $\displaystyle \mV_A,\displaystyle \mV_{base}$ stand for the parameter vector of the model being tuned and that of the base model, respectively. 

\hyperref[fig:cosvsavg]{Figure~\ref*{fig:cosvsavg}} presents the average zero-shot performance on a set of standard benchmarks when $L_A$(PCS) is at different values. The benchmarks include BoolQ~\citep{clark2019boolq}, PIQA~\citep{bisk2020piqa}, HellaSwag~\citep{zellers2019hellaswag}, WinoGrande~\citep{sakaguchi2021winogrande}, ARC-e, ARC-c~\citep{clark2018think}, RACE~\citep{lai2017race} and MMLU~\citep{hendrycks2021measuring}. (c.f. Appendix ~\autoref{tab:detailed performance} for a detailed breakdown of performances on each task.) We can see that despite the original training loss is still present, the model quickly deteriorates to random guesses as the cosine similarity detaches away from that of the base model. 

These observations indicate that it is fairly hard for the model to preserve the base model's performance without keeping a high cosine similarity to it.

\subsection{Deriving the Invariant Terms}\label{Invariant Terms}

Although the vector direction of model parameters is shown to closely stick to its base model, directly comparing the vector direction through cosine similarity requires both models to reveal their parameters, which is unacceptable in many cases. In addition, apart from training, parameter vector direction is vulnerable to some attacks that directly rearrange the model weights. For example, since the hidden units in a model layer are permutation-invariant, one can easily alter the parameter vector direction by randomly permuting the hidden units along with the weights wired to the units. 
\begin{wrapfigure}{l}{0.28\textwidth}
\vspace{-5mm}
\includegraphics[width=0.28\textwidth]{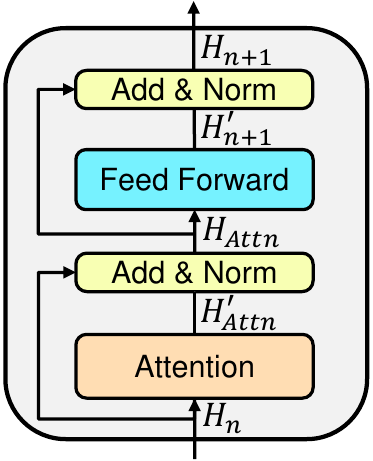}
\caption{Transformer layer}
\label{attention layer}
\vspace{-6mm}
\end{wrapfigure}
These attacks are invisible to discover since they could easily break the cosine similarity but neither change the model structure nor affect the model performance. 

In this subsection, we are going to first systematically analyze and formalize possible weight rearrangements by leveraging the structure constraints of the Transformer, and then derive three terms that are invariant under these rearrangements, even when they are combined.
Let's first consider the Transformer layer as depicted in~\autoref{attention layer}. 
Formally, the layer conducts the following computation:
\begin{equation}\label{eq:attention}
    \displaystyle \mH_{Attn}^{'} = \text{softmax}\left(\frac{\displaystyle \mH_{n}\displaystyle \mW_Q(\displaystyle \mH_{n}\displaystyle \mW_K)^T}{\sqrt{d}}\right)\displaystyle \mH_{n}\displaystyle \mW_V \displaystyle \mW_O
\end{equation}
\begin{equation}\label{eq:ffn}
    \displaystyle \mH_{n+1}^{'} = \sigma\left(\displaystyle \mH_{Attn}\displaystyle \mW_1 + \mathbf{b}_1\right)\displaystyle \mW_2 + \mathbf{b}_2
\end{equation}
where $\displaystyle \mH_{n} \in \mathbb{R}^{l \times d}$ is the hidden state of the $n$-th layer, with $l, d$ being sequence length and model dimensions, respectively. 
$\displaystyle \mH_{Attn}^{'}$ is the self-attention output. To reduce clutter, we omit equations related to residual connection and LayerNorm, but denote the variables right before it with an apostrophe. The $\displaystyle \mW$'s and $\mathbf{b}$'s are weights and biases. 

Note that the first layer reads the word embedding, i.e., $\displaystyle \mH_{0}=\displaystyle \mX \in \mathbb{R}^{l \times d}$, and the final output distribution $\mathbf{P} \in \mathbb{R}^{l \times v}$ is given by 
\begin{equation}\label{eq:finalsoftmax}
    \mathbf{P} = \text{softmax}\left(\displaystyle \mH_{N}\displaystyle \mE\right)
\end{equation}
where $v$ is the vocabulary size, $N$ is the total number of layers, and $\displaystyle \mE \in \mathbb{R}^{d \times v}$ is the parameter matrix in the softmax layer, which is sometimes tied with the word embedding matrix at the input.  

\subsubsection{Forms of Weight Rearrangement Attacks}
Putting~\autoref{eq:attention}\textasciitilde~\autoref{eq:finalsoftmax} together, we can systematically analyze how the parameter vector direction can be attacked through direct weight rearrangements. There are totally 3 forms of attacks that could camouflage the model without changing its architecture or affecting its output.

\textbf{1. Linear mapping attack on $\displaystyle \mW_Q, \displaystyle \mW_K$ and $\displaystyle \mW_V, \displaystyle \mW_O$.} Consider~\autoref{eq:attention}, one can transform $\displaystyle \mW_Q$ and $\displaystyle \mW_K$ symmetrically so that the product $\displaystyle \mW_Q \displaystyle \mW_K^T$ remains unchanged but both weights are significantly modified. This will alter the parameter vector direction significantly. Formally, for any invertible matrix $\displaystyle \mC_{1}$, let
\begin{equation}\label{eq:linear_mapping}
    \Tilde{\displaystyle \mW_Q} = \displaystyle \mW_Q\displaystyle \mC_{1}, \quad \Tilde{\displaystyle \mW_K} = \displaystyle \mW_K\displaystyle \mC_{1}^{-1}
\end{equation}
and substitute them respectively into the model, one can camouflage it as if it's a brand new model, without sacrificing any of the base model's performance. The same holds for $\displaystyle \mW_V, \displaystyle \mW_O$ as well. 

\textbf{2. Permutation attack on $\displaystyle \mW_1, \mathbf{b}_1, \displaystyle \mW_2$.} Consider~\autoref{eq:ffn}, since it consists of two fully connected layers, one can randomly permute the hidden states in the middle layer without changing its output. Formally, let $\displaystyle \mP_{FFN}$ be an arbitrary permutation matrix, one can camouflage the model without sacrificing its performance by substituting the following three matrices accordingly
\begin{equation}\label{eq:ffn_permutation}
    \Tilde{\displaystyle \mW_1} = \displaystyle \mW_1\displaystyle \mP_{FFN}, \quad \Tilde{\displaystyle \mW_2} = \displaystyle \mP_{FFN}^{-1}\displaystyle \mW_{2}, \quad \Tilde{\mathbf{b}_1} = \mathbf{b}_1\displaystyle \mP_{FFN}
\end{equation}
\textbf{3. Permutation attack on word embeddings.} In a similar spirit, one can permute the dimensions in the word embedding matrix as well, although it would require all remaining parameters to be permuted accordingly. Formally, let $\displaystyle \mP_{E}$ be an arbitrary permutation matrix that permutes the dimensions in $\displaystyle \mX$ through $\Tilde{\displaystyle \mX}=\displaystyle \mX\displaystyle \mP_E$, due to the existence of the residual connections, the output of all layers have to be permuted in the same way, i.e., $\Tilde{\displaystyle \mH_n}=\displaystyle \mH_n\displaystyle \mP_E$. Note that it's not necessarily the case in the former two types of attacks. This permutation has to be canceled out at the final softmax layer (\autoref{eq:finalsoftmax}), by permuting the dimensions in $\displaystyle \mE$ accordingly, i.e. $\Tilde{\displaystyle \mE}=\displaystyle \mP_E^{-1}\displaystyle \mE$. Specifically, all remaining parameters have to be permuted in the following way: 
\begin{equation}\label{eq:embedding_permutation}
\begin{aligned}
    &\Tilde{\displaystyle \mW_Q} = \displaystyle \mP_{E}^{-1} \displaystyle \mW_Q, \quad \Tilde{\displaystyle \mW_K} = \displaystyle \mP_{E}^{-1} \displaystyle \mW_K, \quad \Tilde{\displaystyle \mW_V} = \displaystyle \mP_{E}^{-1} \displaystyle \mW_V, \quad \Tilde{\displaystyle \mW_O} = \displaystyle \mW_O \displaystyle \mP_{E} \\
    &\Tilde{\displaystyle \mW_1} = \displaystyle \mP_{E}^{-1} \displaystyle \mW_1, \quad \Tilde{\displaystyle \mW_2} = \displaystyle \mW_{2}\displaystyle \mP_{E}, \quad \Tilde{\mathbf{b}_2} = \mathbf{b}_2 \displaystyle \mP_{E} 
\end{aligned}
\end{equation}
Moreover, putting everything together, one can combine all the aforementioned three types of attacks altogether. Formally, the parameters can be camouflaged as:
\begin{equation}\label{eq:altogether}
\begin{aligned}
    &\Tilde{\displaystyle \mW_Q} = \displaystyle \mP_{E}^{-1} \displaystyle \mW_Q\displaystyle \mC_{1}, \quad \Tilde{\displaystyle \mW_K} = \displaystyle \mP_{E}^{-1} \displaystyle \mW_K\displaystyle \mC_{1}^{-T}, \quad \Tilde{\displaystyle \mW_V} = \displaystyle \mP_{E}^{-1} \displaystyle \mW_V\displaystyle \mC_{2}, \quad \Tilde{\displaystyle \mW_O} = \displaystyle \mC_{2}^{-1}\displaystyle \mW_O \displaystyle \mP_{E} \\
    &\Tilde{\displaystyle \mW_1} = \displaystyle \mP_{E}^{-1} \displaystyle \mW_1\displaystyle \mP_{FFN}, \quad \Tilde{\mathbf{b}_1} = \mathbf{b}_1\displaystyle \mP_{FFN}, \quad \Tilde{\displaystyle \mW_2} = \displaystyle \mP_{FFN}^{-1}\displaystyle \mW_{2}\displaystyle \mP_{E}, \quad \Tilde{\mathbf{b}_2} = \mathbf{b}_2 \displaystyle \mP_{E} \\
    &\Tilde{\displaystyle \mX}=\displaystyle \mX\displaystyle \mP_E, \quad \Tilde{\displaystyle \mE}=\displaystyle \mP_E^{-1}\displaystyle \mE
\end{aligned}
\end{equation}
Note that for permutation matrix we have $\displaystyle \mP^{-1}=\displaystyle \mP^{T}$. This includes all possible attacks that 1) do not change the model architecture, and 2) do not affect the model's output. 
\subsubsection{The Invariant Terms to These Attacks}\label{invariant}
To find the invariant terms under all these attacks, we need to combine terms in~\autoref{eq:altogether} to get the invariant term that nicely cancels out all extra camouflaging matrices. To this end, we construct 3 invariant terms:
\begin{equation}\label{eq:invariants}
    \displaystyle \mM_a = \hat{\displaystyle \mX}\displaystyle \mW_Q\displaystyle \mW_K^T\hat{\displaystyle \mX}^{T}, \quad 
    \displaystyle \mM_b = \hat{\displaystyle \mX}\displaystyle \mW_V\displaystyle \mW_O\hat{\displaystyle \mX}^{T}, \quad
    \displaystyle \mM_f = \hat{\displaystyle \mX}\displaystyle \mW_1\displaystyle \mW_2\hat{\displaystyle \mX}^{T}
\end{equation}
Note that for \(\hat{\displaystyle \mX}\)
in these terms, we don't include all tokens from a vocabulary or tokens in a specific sentence; instead, we select a subset of tokens. There are two problems if we directly use all tokens' embeddings $\displaystyle \mX$. First, using the whole embedding matrix will make the terms unnecessarily large and of variable size between different models. Second, more importantly, since it is common to inherit a base model with an augmented vocabulary, i.e., to append a set of new tokens at the end of the original vocabulary, the invariant terms would have different sizes and be incomparable. Third, if we designate specific tokens instead, the selected tokens may not always exist in all LLMs being tested. Consequently, we carefully choose the tokens to be included in $\hat{\displaystyle \mX}$, by following these steps:
\begin{enumerate}[label=\arabic*., nosep,leftmargin=*]
\item Select a sufficiently big corpus as a standard verifying corpus.
\item Tokenize the corpus with the LLM's vocabulary and sort all tokens according to their frequency.
\item Delete all tokens in the vocabulary that don't show up in the corpus.
\item Among the remaining tokens, select the least frequent $K$ tokens as the tokens to be included in $\hat{\displaystyle \mX}$. 
\end{enumerate}
Here, using a standard corpus ensures that the resulting tokenization will be identical if a certain model's vocabulary is a subset of another; the sufficiently large corpus stabilizes the frequencies of tokens in the vocabulary and provides enough chance for as many tokens as possible to show up. Deleting zero-shot tokens automatically sweeps off augmented tokens. Selecting the rarest tokens minimizes potential affections brought by parameter updates in subsequent training processes. A properly large $K$ will ensure a large enough set of tokens is included, making the resulting invariant terms more generally representative. More importantly, it will make all the invariant terms have the same size across all LLMs, regardless of their original sizes, i.e., $\displaystyle \mM_a, \displaystyle \mM_b, \displaystyle \mM_f \in \mathbb{R}^{K \times K}$, regardless of the index of the layer or LLM sizes. 

As a result, we can tile up them to form a 3D input tensor $\displaystyle \mM \in \mathbb{R}^{K \times K \times C}$, where $C$ is the channel dimension. If we are using all layers, $C=3N$. Again, in order to make $\displaystyle \mM$ the same size across all models, we only involve the last $r$ layers in the LLM\footnote{In fact, experimentally we find that a small $r$ is already sufficient to discriminate LLMs, it's not necessary to involve many layers. In all of our experiments, $r=2$, so there are only $6$ channels in the input.}.
We show the cosine similarity between the invariant terms in~\autoref{normal cosine similarities},
they still preserve a high correlation to the base model.
\vspace{-2mm}
\section{Mapping the Invariant Terms to Image and Publish it through ZKP}\label{The Fingerprinting Model}

Although invariant terms serve as robust and effective representations for LLMs, we cannot directly publish them due to the potential risk of leaking hidden information, including the size, statistical features, and distribution of parameters. Therefore, we further process invariant terms by mapping them into an image through the fingerprinting model and then publish the image fingerprint instead. This approach helps mitigate the risk of leakage while providing a human-readable fingerprint. 

\begin{figure*}[t]
\centering
\includegraphics[width=\textwidth]{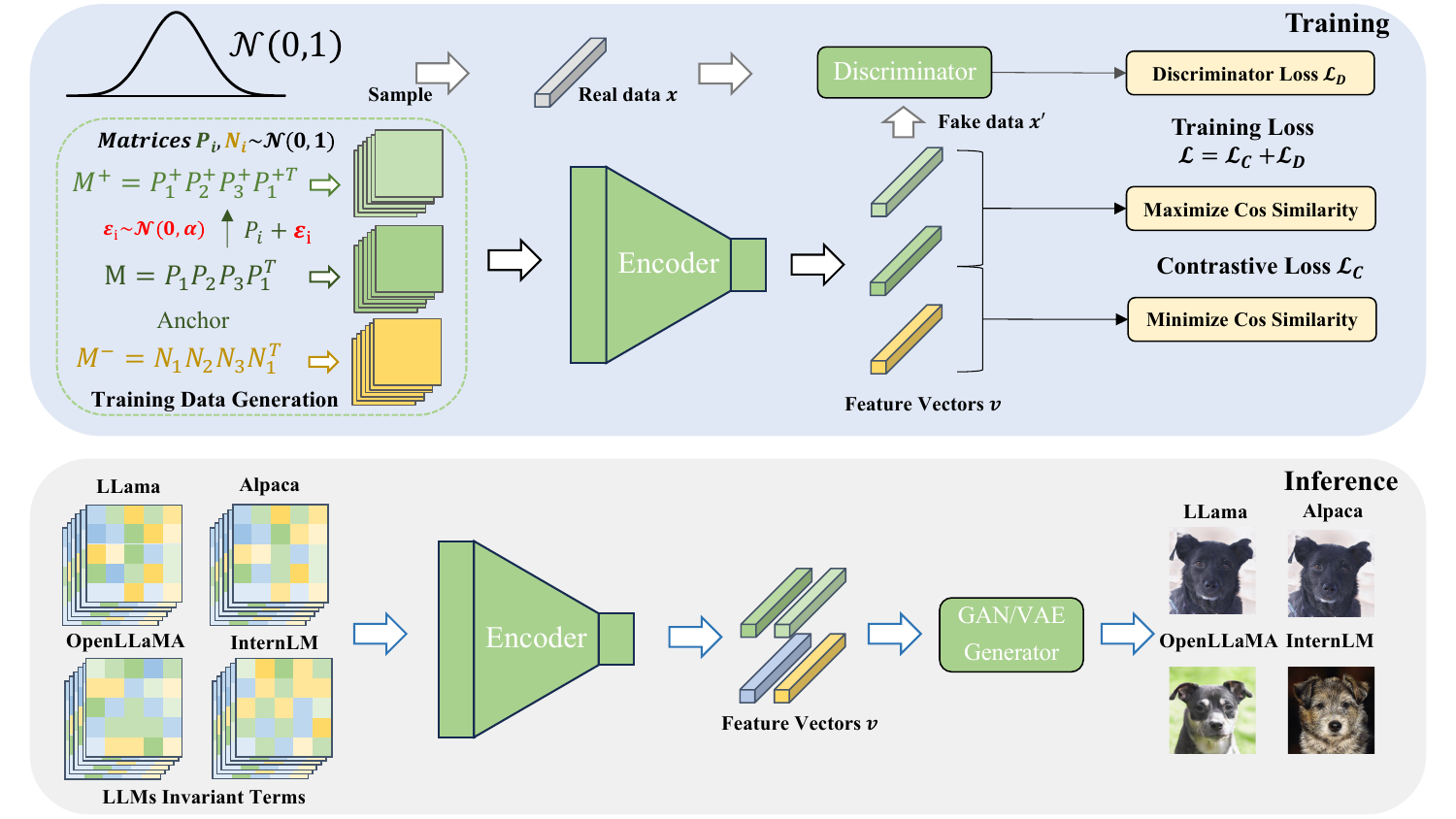} 
\caption{The training and inference of our fingerprinting model.}
\label{fig: model_arch}
\end{figure*}

The fingerprinting model consists of a neural network encoder and an off-the-shelf image generator as depicted in~\autoref{fig: model_arch}. In principle, the encoder takes as input the invariant terms of a certain model, tile them together, and deterministically maps them to a vector that appears to be filled with Gaussian variables. The subsequent image generator reads this vector and maps it to a natural image. Importantly, throughout the process, the locality of the inputs has to be preserved from end to end. i.e., similar invariable terms should result in similar Gaussian variables and finally similar images.

\vspace{-2mm}
\subsection{Training }
The encoder is the only component that needs to be trained in our fingerprinting model. Note that we don't need to use any real LLM weights for training the encoder (all the training data are synthesized by randomly sampled matrices), as it only needs to learn a locality-preserving mapping between the input tensor and the output Gaussian vector.  We adopt contrastive learning to learn locality-preserving mapping. To render the output vector to be Gaussian, we adopt the standard GAN~\citep{karras2019stylebased} training scheme. (c.f.~\autoref{Details of Data Synthesis and Encoder training} for details of data synthesis and the whole training process.)

\subsection{Inference}

In the inference stage, the encoder takes the invariant terms from real LLMs and outputs $\displaystyle \vv$. One image generator converts $\displaystyle \vv$ into a natural image. In principle, any image generator that takes a Gaussian input and has the locality-preserving property would fit here. By visually representing the invariant terms as fingerprints, we can easily identify base models based on their fingerprint images, enabling reliable tracking of model origins. In this paper, we employ the StyleGAN2 generator pretrained on the AFHQ~\citep{choi2020stargan} dog dataset to generate natural images, we detailed it in~\appendixref{StyleGAN2 Generator}.

\subsection{Zero-knowledge Proof for Fingerprints}
In our black-box setting, users are unable to access the model parameters, which presents a significant challenge in ensuring the fingerprint is genuinely derived from the claimed LLM parameters. To address this issue, we employ zero-knowledge proof, a cryptographic technique that allows the prover to convince the verifier that a statement is true without revealing any information beyond the statement’s validity~\citep{ben2013snarks,goldwasser2019knowledge,ChiesaHMMVW20}.

The manufacturer generates a publicly verifiable zero-knowledge proof along with computing the fingerprint, ensuring two critical aspects: (1) the input parameters indeed originate from the specific LLM the manufacturer claims, thereby safeguarding against substitution attack. (c.f.~\autoref{Substitution Attack} for detailed discussion of substitution attack.) (2) the human-readable fingerprint is calculated correctly, confirming the fingerprint is genuinely derived from the LLM parameters. The detailed zero-knowledge proof generation process is as follows:
\begin{enumerate}[label=\arabic*., nosep,leftmargin=*]
    \item Select a random number $t$, commit to LLM parameters (which we denote by $model$) and input $\hat{\mX}$,  $\operatorname{commit}(model,\hat{\mX},t)=\mathbf{cm}$. The commitment $\mathbf{cm}$ is public and does not reveal any information about the model.
    \label{a}
    \item While calculating fingerprint, generate a ZK proof $\pi_1$ prove that the manufacturer knows $model, \hat{\mX}, t$ s.t.
    \begin{enumerate}[ nosep,leftmargin=*]
        \item  $model$ 
        is the claimed LLM parameters 
        and $\hat{\mX}$
        satisfy $\operatorname{commit}(model, \hat{\mX}, t)=\mathbf{cm}$;
        \label{2a}
        \item \label{2b}
        The last two layers parameters $\mW_Q,\mW_K,\mW_V,\mW_O,\mW_1,\mW_2$ in $model$ and input $\hat{\mX}$ satisfy 
        \begin{equation}\label{eq:invariants2}
        \displaystyle \mM_a = \hat{\displaystyle \mX}\displaystyle \mW_Q\displaystyle \mW_K^T\hat{\displaystyle \mX}^{T}, \quad 
        \displaystyle \mM_b = \hat{\displaystyle \mX}\displaystyle \mW_V\displaystyle \mW_O\hat{\displaystyle \mX}^{T}, \quad
        \displaystyle \mM_f = \hat{\displaystyle \mX}\displaystyle \mW_1\displaystyle \mW_2\hat{\displaystyle \mX}^{T}
    \end{equation}
    \item The output human-readable fingerprint is indeed calculated from the invariant terms above.
    \label{2c}
    \end{enumerate}
\end{enumerate}

As above(~\autoref{a} and \autoref{2a}), the manufacturers commit to the claimed model and publish the commitment first, which is a conventional approach to ensure the parameters are not altered during the proof generation. All subsequent proof and inference processes will be carried out with this commitment, and anyone can verify if the model parameters used match those sealed within the commitment. The steps \autoref{2b} and \autoref{2c} are to ensure that the invariant terms and fingerprint are correctly calculated. Anyone who gets proof $\pi_1$ and the commitment $\mathbf{cm}$ can verify that the fingerprint is calculated based on LLM. 

Moreover, we also provide a limited quantitative comparison scheme, which supports one-to-one comparison with open-source models. The manufacturers calculate the cosine similarity of invariant terms and give the zero-knowledge proof $\pi_2$ of this calculation process. Anyone who gets the proof $\pi_2$ and the open-source model can verify the cosine similarity without learning the private model. 


\vspace{-2mm}
\section{Experiments}\label{experiments}
Our experiment is twofold. First, we validated the effectiveness and robustness of invariant terms for identifying the base model. Second, we generated fingerprints based on invariant terms for 80 LLMs and quantitatively assessed their discrimination ability through a human subject study.
\subsection{Effectiveness and Robustness of Invariant Terms}
In this subsection, we validate the effectiveness and robustness of invariant terms in identifying base model through four key experiments. First, we compute the \textbf{I}nvariant Terms' \textbf{C}osine \textbf{S}imilarity (ICS) between 8 widely used open-sourced LLM base models and their offspring models (including heavily continue-pretrained models), verifying its robustness against subsequent training processes. Second, we conduct extensive experiments on additional open-sourced LLMs, showcasing low ICS between 28 independently trained models. Third, we gather 51 offspring models and calculate the accuracy of correctly identifying the base model. Finally, we compare our methods with two latest baselines. 

\subsubsection{High ICS between Base LLMs and Their Offspring Models}\label{sec:fingerprints1}
\begin{table*}[t!]
\centering 
\small 
\setlength\tabcolsep{3pt} 
\begin{tabular}{@{}l|ccccccc@{}}
\toprule
\textbf{ICS} & Falcon-40B  & LLaMA2-13B  & MPT-30B            & LLaMA2-7B            & Qwen-7B              & Baichuan-13B             & InternLM-7B            \\
\midrule
Offspring1 & 99.61       & 99.50       & 99.99       & 99.47      & 98.98 & 99.76 & 99.28            \\
Offspring2 & 99.69       & 99.49       & 99.99       & 99.41      & 99.71   & 99.98 & 99.02    \\
\bottomrule
\end{tabular}
\caption{The ICS between offspring models and their corresponding base model.}
\label{offspring models' ics}
\end{table*}

First, we perform experiments on 7 commonly used open-sourced LLMs, ranging in size from 7B to 40B. The 7 base models considered are Falcon-40B~\citep{almazrouei2023falcon}, MPT-30B~\citep{lin2023mpt}, LLaMA2-7B, 13B, Qwen-7B~\citep{Qwen-7B}, Internlm-7B and Baichuan-13B. For each of these base models, we collect 2 popular offspring models. We extract the invariant terms for all these models and calculate the ICS for each offspring model w.r.t. its base model (\autoref{offspring models' ics}). Remarkably, all offspring models exhibit very high ICS, with an \textbf{average ICS} of $\mathbf{99.56}$.

Second, we leverage the LLaMA-7B base model as a testing ground to assess the robustness of invariant terms under diverse subsequent training processes. We include 10 offspring models detailed in~\autoref{Sufficiency3.1} and add Beaver, Guanaco~\citep{joseph_cheung_2023}, and BiLLa~\citep{BiLLA} to the collection. See Appendix~\autoref{tab:all offspring model descriptions} for detailed descriptions. We extract invariant terms following the previous settings and compute the cosine similarity of the invariant terms (ICS) between each pair of models. Despite undergoing various training paradigms, such as RLHF, SFT, modality extension, and continued pretraining in a new language, we observe a high degree of similarity (\autoref{llama family}), with an \textbf{average ICS} of $\mathbf{94.14}$.
\begin{table*}[t!]
\scriptsize 
\centering
\setlength\tabcolsep{2.1pt} 
\begin{tabular}{@{}l|cccccccccccccc@{}}
\toprule
\textbf{ICS} & \model & MiGPT & Alpaca & MAlpaca & Vicuna & Wizard & Baize & AlpacaL & CAlpaca & Koala & C\model & Beaver & Guanaco & BiLLa \\
\midrule
\model & 100.00 & 99.20 & 99.95 & 99.86 & 99.42 & 99.89 & 99.60 & 99.60 & 91.35 & 99.63 & 93.57 & 99.97 & 92.62 & 82.56 \\
MiGPT & 99.20 & 100.00 & 99.17 & 99.10 & 99.10 & 99.15 & 98.83 & 98.82 & 90.65 & 99.00 & 92.84 & 99.19 & 91.93 & 82.24 \\
Alpaca & 99.95 & 99.17 & 100.00 & 99.82 & 99.38 & 99.85 & 99.55 & 99.57 & 91.31 & 99.59 & 93.53 & 99.97 & 92.59 & 82.52 \\
MAlpaca & 99.86 & 99.10 & 99.82 & 100.00 & 99.31 & 99.76 & 99.46 & 99.47 & 91.23 & 99.51 & 93.45 & 99.84 & 92.50 & 82.51 \\
Vicuna & 99.42 & 99.10 & 99.38 & 99.31 & 100.00 & 99.35 & 99.05 & 99.04 & 90.84 & 99.15 & 93.04 & 99.41 & 92.14 & 82.28 \\
Wizard & 99.89 & 99.15 & 99.85 & 99.76 & 99.35 & 100.00 & 99.50 & 99.50 & 91.25 & 99.56 & 93.47 & 99.87 & 92.52 & 82.57 \\
Baize & 99.60 & 98.83 & 99.55 & 99.46 & 99.05 & 99.50 & 100.00 & 99.23 & 90.97 & 99.25 & 93.19 & 99.57 & 92.25 & 82.25 \\
AlpacaL & 99.60 & 98.82 & 99.57 & 99.47 & 99.04 & 99.50 & 99.23 & 100.00 & 90.99 & 99.24 & 93.21 & 99.59 & 92.31 & 82.30 \\
CAlpaca & 91.35 & 90.65 & 91.31 & 91.23 & 90.84 & 91.25 & 90.97 & 90.99 & 100.00 & 91.04 & 97.44 & 91.33 & 85.19 & 75.60 \\
Koala & 99.63 & 99.00 & 99.59 & 99.51 & 99.15 & 99.56 & 99.25 & 99.24 & 91.04 & 100.00 & 93.23 & 99.61 & 92.27 & 82.34 \\
C\model & 93.57 & 92.84 & 93.53 & 93.45 & 93.04 & 93.47 & 93.19 & 93.21 & 97.44 & 93.23 & 100.00 & 93.55 & 86.80 & 77.41 \\
Beaver & 99.97 & 99.19 & 99.97 & 99.84 & 99.41 & 99.87 & 99.57 & 99.59 & 91.33 & 99.61 & 93.55 & 100.00 & 92.60 & 82.57 \\
Guanaco & 92.62 & 91.93 & 92.59 & 92.50 & 92.14 & 92.52 & 92.25 & 92.31 & 85.19 & 92.27 & 86.80 & 92.60 & 100.00 & 77.17 \\
BiLLa & 82.56 & 82.24 & 82.52 & 82.51 & 82.28 & 82.57 & 82.25 & 82.30 & 75.60 & 82.34 & 77.41 & 82.57 & 77.17 & 100.00 \\
\bottomrule
\end{tabular}
\caption{ The cosine similarities of invariant terms (ICS) between various pairs of \model-7B and its offspring models. Abbreviations: MedAlpaca (MAlpaca), Alpaca-Lora (AlpacaL), MiniGPT-4 (MiGPT), WizardLM (Wizard), Chinese-Alpaca (CAlpaca), Chinese-LLaMA (CLLaMA).}
\vspace{-3mm}
\label{llama family}
\end{table*}

\subsubsection{Low ICS between 28 Independently Trained LLMs}\label{Diverse Set of Independently Trained LLMs}
Besides the aforementioned base models, we assemble a comprehensive collection of 28 open-sourced LLMs, ranging in size from 774M (GPT2-Large) to 180B (Falcon-180B). Please refer to~\appendixref{open-sourced LLMs} for details. We extract invariant terms and calculate ICS between each pair of models. Notably, the similarities between different models were consistently low, with an \textbf{average ICS} of $\mathbf{0.38}$, affirming the effectiveness of invariant terms. (c.f. Appendix~\autoref{tab:28 LLMs} for detailed ICSs.)

\subsubsection{Accuracy in Identify 51 Offspring Models' Base Model}\label{Accuracy in Identify 51 Offspring Models' Base Model}
To assess the effectiveness of our method, we gathered 51 offspring models derived from 18 distinct base models. (c.f.
~\autoref{tab:all offspring model descriptions}
for detailed list and description.) Calculating the ICS between each offspring model and the 18 base models, we predicted the base model with the highest ICS. Comparing these predictions with the ground truth, our method accurately identified the base models for all 51 offspring LLMs, achieving $\mathbf{100\%}$ \textbf{accuracy}.

\begin{table*}[t!]
\scriptsize 
\centering
\setlength\tabcolsep{2.7pt} 
\begin{tabular}{@{}l|ccccccccccccccc@{}}
\toprule
\textbf{FSR}  & MiGPT & Alpaca & MAlpaca & Vicuna & Wizard & Baize & AlpacaL & CAlpaca & Koala & C\model & Beaver & Guanaco & BiLLa & Avg. \\
\midrule
Trap & 0 & 8.04 & 24.14 & 2.30 & 17.24 & 44.83 & 39.08 & 1.15 & 0 & 0 & 8.05 & 10.34& 0 & \textbf{11.94} \\
IF$_\text{adapter}^1$ & 0 & 10 & 0 & 0 & 0 & 40 & 10 & 0 & 0 & 0 & 10 & 0 & 0 & \textbf{5.38} \\
IF$_\text{adapter}^2$ & 100 & 100 & 100 & 100 & 100 & 100 & 100 & 20 & 100 & 50 & 100 & 30 & 100 & \textbf{84.62} \\
Ours & 100 & 100 & 100 & 100 & 100 & 100 & 100 & 100 & 100 & 100 & 100 & 100 & 100 & \textbf{100.00} \\
\bottomrule
\end{tabular}
\caption{ Different methods' FSR on various \model's offspring models. IF$_\text{adapter}^1$ and IF$_\text{adapter}^2$ represent two different experimental settings of IF, with the former using all parameters and the latter only using the embedding parameter. Abbreviations are consistent with~\autoref{llama family}.}
\label{two baselines}
\end{table*}

\begin{figure*}[t!]
\centering
\includegraphics[width=\textwidth]{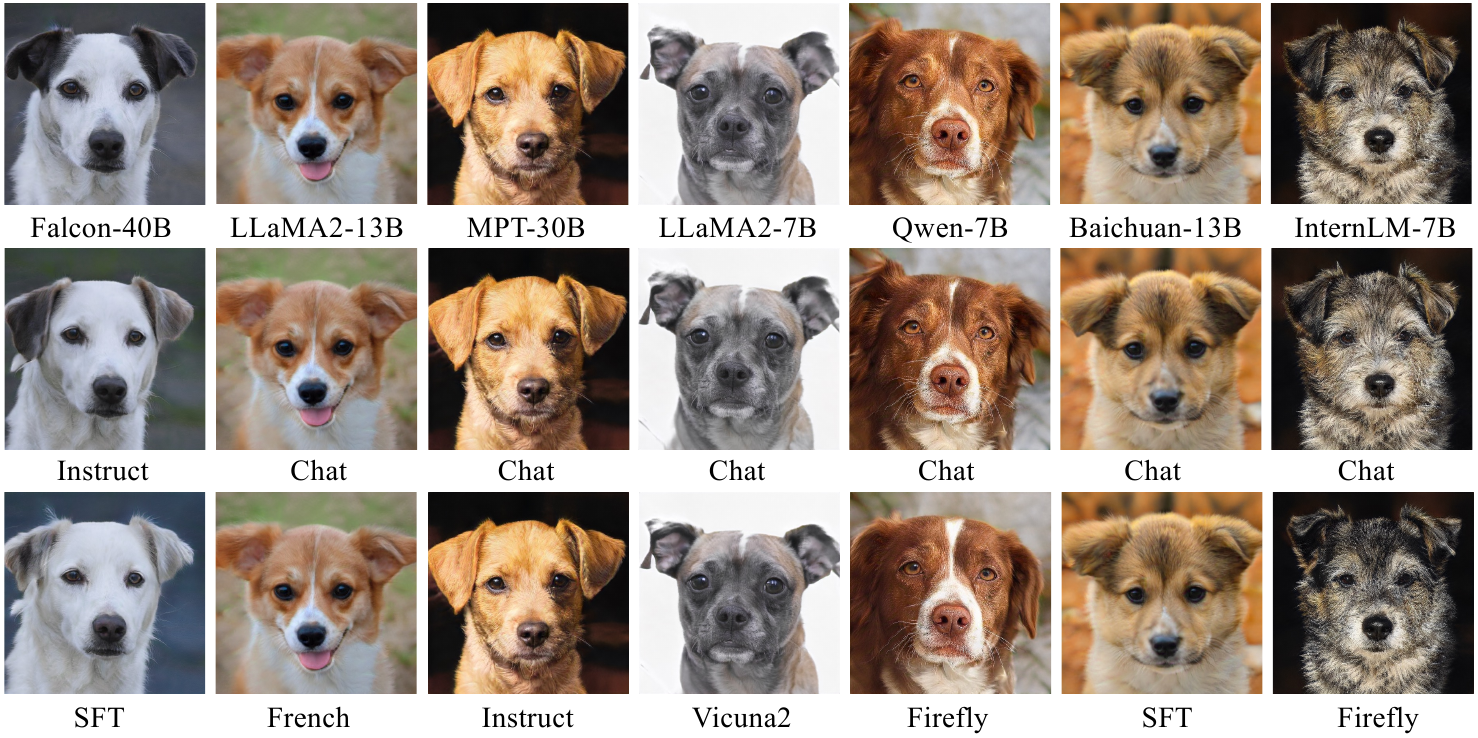} 
\caption{Fingerprints of 7 different base models (in the first row) and their corresponding offspring models (the lower two rows) are presented. The base model's name is omitted in the offspring models. }
\vspace{-3mm}
\label{7 base models and their corresponding models}
\end{figure*}
\vspace{-3mm}
\subsubsection{Comparing to Latest Fingerprinting Methods}\label{Comparing to Latest Fingerprinting Methods}
\vspace{-2mm}

There are few fingerprinting methods designed for LLMs. Trap~\citep{gubri2024trap} optimizes adversarial suffixes to elicit specific responses, while IF~\citep{xu2024instructional} fintuned LLMs to make them generate predefined answers. We tested the Fingerprint Success Rate (FSR) \citep{gu2023watermarking} of these methods on \model's offspring models. Our method demonstrates superior performance (\autoref{two baselines}), even when compared to the white-box method IF$_\text{adapter}$. (More illustrations in~\autoref{More Illustrations of Baseline Comparison}.)
\vspace{-2mm}
\subsection{Discrimination Ability of Human-readable Fingerprints }
\vspace{-2mm}


Based on previous invariant terms, we employ the fingerprinting model illustrated in~\autoref{The Fingerprinting Model} to generate and publish human-readable fingerprints for previously mentioned LLMs. In \autoref{7 base models and their corresponding models}, there are fingerprints of the 7 independently trained LLMs and their offspring models (\autoref{sec:fingerprints1}). Notably, for all the offspring models, their fingerprints closely resemble those of their base models. On the other hand, LLMs based on different models yield highly distinctive fingerprints, encompassing various appearances and breeds of dogs. Due to space limit, the fingerprints of LLaMA family models, the rest offspring models, and the 28 independently trained LLMs are listed in \appendixref{more fingerprints}.

Furthermore, we conducted a human subject study and yielded a $\mathbf{94.74\%}$ \textbf{accuracy} rate (c.f.~\appendixref{Human Subject Study} for details), quantitatively demonstrate the discrimination ability of our generated fingerprints. Although using human-readable fingerprints introduces minor losses, manufacturers can provide one-to-one comparison results with proof to make up for this loss of misjudgment.

Except for the aforementioned experiments, we independently train LLMs on a smaller scale to provide further validation for our method. (c.f.~\appendixref{Additional  Experiments})
\vspace{-3mm}
\section{Conclusion}
\vspace{-2mm}
In this paper, we introduce a novel approach that generates a human-readable fingerprint for LLM. Owing to Zero-Knowledge Proof, all fingerprinting steps are internally conducted by the LLMs owners. Our method is actually a black-box method as only the image fingerprint and corresponding proof need to be released. There is no exposure of model weights or information leakage to the public throughout the entire process.
Furthermore, we detailed our works' limitations in~\autoref{Limitations}.

\section{Acknowledgements}
We would like to thank Shiyu Liang, Siyuan Huang, and the anonymous reviewers for helpful discussions and feedback. This work was sponsored by the National Key Research and Development Program of China (No. 2023ZD0121402) and National Natural Science Foundation of China (NSFC) grant (No.62106143).


\bibliographystyle{nips2024}
\bibliography{nips2024}

\begin{thebibliography}{103}
\providecommand{\natexlab}[1]{#1}
\providecommand{\url}[1]{\texttt{#1}}
\expandafter\ifx\csname urlstyle\endcsname\relax
  \providecommand{\doi}[1]{doi: #1}\else
  \providecommand{\doi}{doi: \begingroup \urlstyle{rm}\Url}\fi

\bibitem[Abdelnabi \& Fritz(2021)Abdelnabi and Fritz]{nlp_abdelnabi2021awl}
Abdelnabi, S. and Fritz, M.
\newblock Adversarial watermarking transformer: Towards tracing text provenance with data hiding.
\newblock In \emph{IEEE Symposium on Security and Privacy (S\&P)}, pp.\  121--140, 2021.

\bibitem[Adi et~al.(2018{\natexlab{a}})Adi, Baum, Cisse, Pinkas, and Keshet]{adi2018turning}
Adi, Y., Baum, C., Cisse, M., Pinkas, B., and Keshet, J.
\newblock Turning your weakness into a strength: Watermarking deep neural networks by backdooring.
\newblock In \emph{27th USENIX Security Symposium (USENIX Security 18)}, pp.\  1615--1631, 2018{\natexlab{a}}.

\bibitem[Adi et~al.(2018{\natexlab{b}})Adi, Baum, Cisse, Pinkas, and Keshet]{black_adi2018turning}
Adi, Y., Baum, C., Cisse, M., Pinkas, B., and Keshet, J.
\newblock Turning your weakness into a strength: Watermarking deep neural networks by backdooring.
\newblock In \emph{27th USENIX Security Symposium (USENIX Security 18)}, pp.\  1615--1631, 2018{\natexlab{b}}.

\bibitem[Almazrouei et~al.(2023)Almazrouei, Alobeidli, Alshamsi, Cappelli, Cojocaru, Debbah, Goffinet, Hesslow, Launay, Malartic, et~al.]{almazrouei2023falcon}
Almazrouei, E., Alobeidli, H., Alshamsi, A., Cappelli, A., Cojocaru, R., Debbah, M., Goffinet, {\'E}., Hesslow, D., Launay, J., Malartic, Q., et~al.
\newblock The falcon series of open language models.
\newblock \emph{arXiv preprint arXiv:2311.16867}, 2023.

\bibitem[Bai et~al.(2023)Bai, Bai, Chu, Cui, Dang, et~al.]{Qwen-7B}
Bai, J., Bai, S., Chu, Y., Cui, Z., Dang, K., et~al.
\newblock Qwen technical report.
\newblock \emph{arXiv preprint arXiv:2309.16609}, 2023.

\bibitem[Bai et~al.(2022)Bai, Jones, Ndousse, Askell, Chen, DasSarma, Drain, Fort, Ganguli, Henighan, et~al.]{bai2022training}
Bai, Y., Jones, A., Ndousse, K., Askell, A., Chen, A., DasSarma, N., Drain, D., Fort, S., Ganguli, D., Henighan, T., et~al.
\newblock Training a helpful and harmless assistant with reinforcement learning from human feedback.
\newblock \emph{arXiv preprint arXiv:2204.05862}, 2022.

\bibitem[BaiChuan-Inc(2023)]{baichuan}
BaiChuan-Inc.
\newblock \url{https://github.com/baichuan-inc/Baichuan-7B}, 2023.

\bibitem[Ben-Sasson et~al.(2013)Ben-Sasson, Chiesa, Genkin, Tromer, and Virza]{ben2013snarks}
Ben-Sasson, E., Chiesa, A., Genkin, D., Tromer, E., and Virza, M.
\newblock Snarks for c: Verifying program executions succinctly and in zero knowledge.
\newblock In \emph{Annual cryptology conference}. Springer, 2013.

\bibitem[Biderman et~al.(2023)Biderman, Schoelkopf, Anthony, Bradley, O’Brien, Hallahan, Khan, Purohit, Prashanth, Raff, et~al.]{biderman2023pythia}
Biderman, S., Schoelkopf, H., Anthony, Q.~G., Bradley, H., O’Brien, K., Hallahan, E., Khan, M.~A., Purohit, S., Prashanth, U.~S., Raff, E., et~al.
\newblock Pythia: A suite for analyzing large language models across training and scaling.
\newblock In \emph{International Conference on Machine Learning}, pp.\  2397--2430. PMLR, 2023.

\bibitem[Bisk et~al.(2020)Bisk, Zellers, Gao, Choi, et~al.]{bisk2020piqa}
Bisk, Y., Zellers, R., Gao, J., Choi, Y., et~al.
\newblock Piqa: Reasoning about physical commonsense in natural language.
\newblock In \emph{Proceedings of the AAAI conference on artificial intelligence}, pp.\  7432--7439, 2020.

\bibitem[Black et~al.(2022)Black, Biderman, Hallahan, Anthony, Gao, Golding, He, Leahy, McDonell, Phang, et~al.]{GPT-NeoX-20}
Black, S., Biderman, S., Hallahan, E., Anthony, Q., Gao, L., Golding, L., He, H., Leahy, C., McDonell, K., Phang, J., et~al.
\newblock Gpt-neox-20b: An open-source autoregressive language model.
\newblock In \emph{Proceedings of BigScience Episode\# 5--Workshop on Challenges \& Perspectives in Creating Large Language Models}, pp.\  95--136, 2022.

\bibitem[Boenisch(2021)]{boenisch2021systematic}
Boenisch, F.
\newblock A systematic review on model watermarking for neural networks.
\newblock \emph{Frontiers in big Data}, 4:\penalty0 729663, 2021.

\bibitem[Brown et~al.(2020)Brown, Mann, Ryder, Subbiah, Kaplan, Dhariwal, Neelakantan, Shyam, Sastry, Askell, et~al.]{brown2020language}
Brown, T., Mann, B., Ryder, N., Subbiah, M., Kaplan, J.~D., Dhariwal, P., Neelakantan, A., Shyam, P., Sastry, G., Askell, A., et~al.
\newblock Language models are few-shot learners.
\newblock \emph{Advances in neural information processing systems}, 33:\penalty0 1877--1901, 2020.

\bibitem[Chen et~al.(2019{\natexlab{a}})Chen, Rouhani, Fu, Zhao, and Koushanfar]{chen2019deepmarks}
Chen, H., Rouhani, B.~D., Fu, C., Zhao, J., and Koushanfar, F.
\newblock Deepmarks: A secure fingerprinting framework for digital rights management of deep learning models.
\newblock In \emph{Proceedings of the 2019 on International Conference on Multimedia Retrieval}, pp.\  105--113, 2019{\natexlab{a}}.

\bibitem[Chen et~al.(2019{\natexlab{b}})Chen, Rouhani, and Koushanfar]{black_chen2019blackmarks}
Chen, H., Rouhani, B.~D., and Koushanfar, F.
\newblock Blackmarks: Blackbox multibit watermarking for deep neural networks.
\newblock \emph{arXiv preprint arXiv:1904.00344}, 2019{\natexlab{b}}.

\bibitem[Chen et~al.(2022)Chen, Zhou, Zhang, Chen, Zhang, Chen, Hua, and Yu]{fingerprinting_chen2022perceptualhash}
Chen, H., Zhou, H., Zhang, J., Chen, D., Zhang, W., Chen, K., Hua, G., and Yu, N.
\newblock Perceptual hashing of deep convolutional neural networks for model copy detection.
\newblock \emph{ACM Transactions on Multimedia Computing, Communications and Applications ({TOMCCAP})}, 2022.

\bibitem[Chen et~al.(2021)Chen, Chen, Zhang, and Wang]{lottery_chen2021you}
Chen, X., Chen, T., Zhang, Z., and Wang, Z.
\newblock You are caught stealing my winning lottery ticket! making a lottery ticket claim its ownership.
\newblock \emph{Advances in Neural Information Processing Systems (NeurIPS)}, 34:\penalty0 1780--1791, 2021.

\bibitem[Chiesa et~al.(2020)Chiesa, Hu, Maller, Mishra, Vesely, and Ward]{ChiesaHMMVW20}
Chiesa, A., Hu, Y., Maller, M., Mishra, P., Vesely, P., and Ward, N.~P.
\newblock Marlin: Preprocessing zksnarks with universal and updatable {SRS}.
\newblock In \emph{39th Annual International Conference on the Theory and Applications of Cryptographic Techniques}. Springer, 2020.

\bibitem[Choi et~al.(2020)Choi, Uh, Yoo, and Ha]{choi2020stargan}
Choi, Y., Uh, Y., Yoo, J., and Ha, J.-W.
\newblock Stargan v2: Diverse image synthesis for multiple domains.
\newblock In \emph{Proceedings of the IEEE/CVF conference on computer vision and pattern recognition}, pp.\  8188--8197, 2020.

\bibitem[Chowdhery et~al.(2022)Chowdhery, Narang, Devlin, Bosma, Mishra, Roberts, Barham, Chung, Sutton, Gehrmann, et~al.]{chowdhery2022palm}
Chowdhery, A., Narang, S., Devlin, J., Bosma, M., Mishra, G., Roberts, A., Barham, P., Chung, H.~W., Sutton, C., Gehrmann, S., et~al.
\newblock Palm: Scaling language modeling with pathways.
\newblock \emph{arXiv preprint arXiv:2204.02311}, 2022.

\bibitem[Christ et~al.(2023)Christ, Gunn, and Zamir]{cryptoeprint:2023/763}
Christ, M., Gunn, S., and Zamir, O.
\newblock Undetectable watermarks for language models.
\newblock \emph{arXiv preprint arXiv:2306.09194}, 2023.

\bibitem[Clark et~al.(2019)Clark, Lee, Chang, Kwiatkowski, Collins, and Toutanova]{clark2019boolq}
Clark, C., Lee, K., Chang, M.-W., Kwiatkowski, T., Collins, M., and Toutanova, K.
\newblock Boolq: Exploring the surprising difficulty of natural yes/no questions.
\newblock \emph{arXiv preprint arXiv:1905.10044}, 2019.

\bibitem[Clark et~al.(2018)Clark, Cowhey, Etzioni, Khot, Sabharwal, Schoenick, and Tafjord]{clark2018think}
Clark, P., Cowhey, I., Etzioni, O., Khot, T., Sabharwal, A., Schoenick, C., and Tafjord, O.
\newblock Think you have solved question answering? try arc, the ai2 reasoning challenge.
\newblock \emph{arXiv preprint arXiv:1803.05457}, 2018.

\bibitem[Computer(2023)]{together2023redpajama}
Computer, T.
\newblock Redpajama: An open source recipe to reproduce llama training dataset, 2023.
\newblock URL \url{https://github.com/togethercomputer/RedPajama-Data}.

\bibitem[Conover et~al.(2023)Conover, Hayes, Mathur, Xie, Wan, Shah, Ghodsi, Wendell, Zaharia, and Xin]{DatabricksBlog2023DollyV2}
Conover, M., Hayes, M., Mathur, A., Xie, J., Wan, J., Shah, S., Ghodsi, A., Wendell, P., Zaharia, M., and Xin, R.
\newblock Free dolly: Introducing the world's first truly open instruction-tuned llm, 2023.
\newblock URL \url{https://www.databricks.com/blog/2023/04/12/dolly-first-open-commercially-viable-instruction-tuned-llm}.

\bibitem[Cui et~al.(2023)Cui, Yang, and Yao]{chinese-llama-alpaca}
Cui, Y., Yang, Z., and Yao, X.
\newblock Efficient and effective text encoding for chinese llama and alpaca.
\newblock \emph{arXiv preprint arXiv:2304.08177}, 2023.

\bibitem[Dettmers et~al.(2023)Dettmers, Pagnoni, Holtzman, and Zettlemoyer]{joseph_cheung_2023}
Dettmers, T., Pagnoni, A., Holtzman, A., and Zettlemoyer, L.
\newblock Qlora: Efficient finetuning of quantized llms.
\newblock \emph{arXiv preprint arXiv:2305.14314}, 2023.

\bibitem[Dey et~al.(2023)Dey, Gosal, Khachane, Marshall, Pathria, Tom, Hestness, et~al.]{dey2023cerebrasgpt}
Dey, N., Gosal, G., Khachane, H., Marshall, W., Pathria, R., Tom, M., Hestness, J., et~al.
\newblock Cerebras-gpt: Open compute-optimal language models trained on the cerebras wafer-scale cluster.
\newblock \emph{arXiv preprint arXiv:2304.03208}, 2023.

\bibitem[Du et~al.(2022)Du, Qian, Liu, Ding, Qiu, Yang, and Tang]{du2022glm}
Du, Z., Qian, Y., Liu, X., Ding, M., Qiu, J., Yang, Z., and Tang, J.
\newblock Glm: General language model pretraining with autoregressive blank infilling.
\newblock In \emph{Proceedings of the 60th Annual Meeting of the Association for Computational Linguistics (Volume 1: Long Papers)}, pp.\  320--335, 2022.

\bibitem[Fan et~al.(2019)Fan, Ng, and Chan]{passport_fan2019rethinking}
Fan, L., Ng, K.~W., and Chan, C.~S.
\newblock Rethinking deep neural network ownership verification: Embedding passports to defeat ambiguity attacks.
\newblock \emph{Advances in Neural Information Processing Systems (NeurIPS)}, 32, 2019.

\bibitem[Fan et~al.(2021)Fan, Ng, Chan, and Yang]{passport_fan2021deepip}
Fan, L., Ng, K.~W., Chan, C.~S., and Yang, Q.
\newblock Deepipr: Deep neural network intellectual property protection with passports.
\newblock \emph{IEEE Transactions on Pattern Analysis and Machine Intelligence (TPAMI)}, 2021.

\bibitem[Gao et~al.(2020)Gao, Biderman, Black, Golding, Hoppe, Foster, Phang, He, Thite, Nabeshima, et~al.]{gao2020pile}
Gao, L., Biderman, S., Black, S., Golding, L., Hoppe, T., Foster, C., Phang, J., He, H., Thite, A., Nabeshima, N., et~al.
\newblock The pile: An 800gb dataset of diverse text for language modeling.
\newblock \emph{arXiv preprint arXiv:2101.00027}, 2020.

\bibitem[Geng(2023)]{geng2023easylm}
Geng, X.
\newblock Easylm: A simple and scalable training framework for large language models, 2023.
\newblock URL \url{https://github.com/young-geng/EasyLM}.

\bibitem[Geng \& Liu(2023)Geng and Liu]{openlm2023openllama}
Geng, X. and Liu, H.
\newblock Openllama: An open reproduction of llama, May 2023.
\newblock URL \url{https://github.com/openlm-research/open_llama}.

\bibitem[Goldwasser et~al.(2019)Goldwasser, Micali, and Rackoff]{goldwasser2019knowledge}
Goldwasser, S., Micali, S., and Rackoff, C.
\newblock The knowledge complexity of interactive proof-systems.
\newblock In \emph{Providing sound foundations for cryptography}. 2019.

\bibitem[GPT-4(2023)]{OpenAI2023GPT4TR}
GPT-4, O.
\newblock Gpt-4 technical report.
\newblock \emph{ArXiv}, abs/2303.08774, 2023.

\bibitem[Gu et~al.(2022)Gu, Huang, Zheng, Chang, and Hsieh]{gu2023watermarking}
Gu, C., Huang, C., Zheng, X., Chang, K.-W., and Hsieh, C.-J.
\newblock Watermarking pre-trained language models with backdooring.
\newblock \emph{arXiv preprint arXiv:2210.07543}, 2022.

\bibitem[Gubri et~al.(2024)Gubri, Ulmer, Lee, Yun, and Oh]{gubri2024trap}
Gubri, M., Ulmer, D., Lee, H., Yun, S., and Oh, S.~J.
\newblock Trap: Targeted random adversarial prompt honeypot for black-box identification.
\newblock \emph{arXiv preprint arXiv:2402.12991}, 2024.

\bibitem[Guo \& Potkonjak(2018)Guo and Potkonjak]{black_guo2018watermarking}
Guo, J. and Potkonjak, M.
\newblock Watermarking deep neural networks for embedded systems.
\newblock In \emph{2018 IEEE/ACM International Conference on Computer-Aided Design (ICCAD)}, pp.\  1--8. IEEE, 2018.

\bibitem[Han et~al.(2023)Han, Adams, Papaioannou, Grundmann, Oberhauser, L{\"o}ser, Truhn, and Bressem]{han2023medalpaca}
Han, T., Adams, L.~C., Papaioannou, J.-M., Grundmann, P., Oberhauser, T., L{\"o}ser, A., Truhn, D., and Bressem, K.~K.
\newblock Medalpaca--an open-source collection of medical conversational ai models and training data.
\newblock \emph{arXiv preprint arXiv:2304.08247}, 2023.

\bibitem[He et~al.(2022{\natexlab{a}})He, Xu, Lyu, Wu, and Wang]{nlp_he2022dawnnlp}
He, X., Xu, Q., Lyu, L., Wu, F., and Wang, C.
\newblock Protecting intellectual property of language generation apis with lexical watermark.
\newblock In \emph{Proceedings of the AAAI Conference on Artificial Intelligence (AAAI)}, volume~36, pp.\  10758--10766, 2022{\natexlab{a}}.

\bibitem[He et~al.(2022{\natexlab{b}})He, Xu, Zeng, Lyu, Wu, Li, and Jia]{he2022cater}
He, X., Xu, Q., Zeng, Y., Lyu, L., Wu, F., Li, J., and Jia, R.
\newblock Cater: Intellectual property protection on text generation apis via conditional watermarks.
\newblock \emph{Advances in Neural Information Processing Systems}, 35:\penalty0 5431--5445, 2022{\natexlab{b}}.

\bibitem[Hendrycks et~al.(2020)Hendrycks, Burns, Basart, Zou, Mazeika, Song, and Steinhardt]{hendrycks2021measuring}
Hendrycks, D., Burns, C., Basart, S., Zou, A., Mazeika, M., Song, D., and Steinhardt, J.
\newblock Measuring massive multitask language understanding.
\newblock In \emph{International Conference on Learning Representations}, 2020.

\bibitem[Hoffmann et~al.(2022)Hoffmann, Borgeaud, Mensch, Buchatskaya, Cai, Rutherford, Casas, Hendricks, Welbl, Clark, et~al.]{hoffmann2022training}
Hoffmann, J., Borgeaud, S., Mensch, A., Buchatskaya, E., Cai, T., Rutherford, E., Casas, D. d.~L., Hendricks, L.~A., Welbl, J., Clark, A., et~al.
\newblock Training compute-optimal large language models.
\newblock \emph{arXiv preprint arXiv:2203.15556}, 2022.

\bibitem[Jia et~al.(2021)Jia, Yaghini, Choquette-Choo, Dullerud, Thudi, Chandrasekaran, and Papernot]{fingerprinting_jia2021proof}
Jia, H., Yaghini, M., Choquette-Choo, C.~A., Dullerud, N., Thudi, A., Chandrasekaran, V., and Papernot, N.
\newblock Proof-of-learning: Definitions and practice.
\newblock In \emph{IEEE Symposium on Security and Privacy (S\&P)}, pp.\  1039--1056. IEEE, 2021.

\bibitem[Karras et~al.(2019)Karras, Laine, and Aila]{karras2019stylebased}
Karras, T., Laine, S., and Aila, T.
\newblock A style-based generator architecture for generative adversarial networks.
\newblock In \emph{Proceedings of the IEEE/CVF conference on computer vision and pattern recognition}, pp.\  4401--4410, 2019.

\bibitem[Karras et~al.(2020)Karras, Laine, Aittala, Hellsten, Lehtinen, and Aila]{karras2020analyzing}
Karras, T., Laine, S., Aittala, M., Hellsten, J., Lehtinen, J., and Aila, T.
\newblock Analyzing and improving the image quality of stylegan.
\newblock In \emph{Proceedings of the IEEE/CVF conference on computer vision and pattern recognition}, pp.\  8110--8119, 2020.

\bibitem[Kate et~al.(2010)Kate, Zaverucha, and Goldberg]{kate2010constant}
Kate, A., Zaverucha, G.~M., and Goldberg, I.
\newblock Constant-size commitments to polynomials and their applications.
\newblock In \emph{Advances in Cryptology-ASIACRYPT 2010: 16th International Conference on the Theory and Application of Cryptology and Information Security, Singapore, December 5-9, 2010. Proceedings 16}, pp.\  177--194. Springer, 2010.

\bibitem[Kirchenbauer et~al.(2023)Kirchenbauer, Geiping, Wen, Katz, Miers, and Goldstein]{kirchenbauer2023watermark}
Kirchenbauer, J., Geiping, J., Wen, Y., Katz, J., Miers, I., and Goldstein, T.
\newblock A watermark for large language models.
\newblock In Krause, A., Brunskill, E., Cho, K., Engelhardt, B., Sabato, S., and Scarlett, J. (eds.), \emph{Proceedings of the 40th International Conference on Machine Learning}, volume 202 of \emph{Proceedings of Machine Learning Research}, pp.\  17061--17084. PMLR, 23--29 Jul 2023.
\newblock URL \url{https://proceedings.mlr.press/v202/kirchenbauer23a.html}.

\bibitem[K{\"o}pf et~al.(2023)K{\"o}pf, Kilcher, von R{\"u}tte, Anagnostidis, Tam, Stevens, Barhoum, Duc, Stanley, Nagyfi, et~al.]{kopf2023openassistant}
K{\"o}pf, A., Kilcher, Y., von R{\"u}tte, D., Anagnostidis, S., Tam, Z.-R., Stevens, K., Barhoum, A., Duc, N.~M., Stanley, O., Nagyfi, R., et~al.
\newblock Openassistant conversations--democratizing large language model alignment.
\newblock \emph{arXiv preprint arXiv:2304.07327}, 2023.

\bibitem[Krishna et~al.(2023)Krishna, Song, Karpinska, Wieting, and Iyyer]{krishna2023paraphrasing}
Krishna, K., Song, Y., Karpinska, M., Wieting, J., and Iyyer, M.
\newblock Paraphrasing evades detectors of ai-generated text, but retrieval is an effective defense.
\newblock \emph{arXiv preprint arXiv:2303.13408}, 2023.

\bibitem[Lai et~al.(2017)Lai, Xie, Liu, Yang, and Hovy]{lai2017race}
Lai, G., Xie, Q., Liu, H., Yang, Y., and Hovy, E.
\newblock Race: Large-scale reading comprehension dataset from examinations.
\newblock In \emph{Proceedings of the 2017 Conference on Empirical Methods in Natural Language Processing}, pp.\  785--794, 2017.

\bibitem[Le~Merrer et~al.(2020)Le~Merrer, Perez, and Tr{\'e}dan]{black_le2020adversarial}
Le~Merrer, E., Perez, P., and Tr{\'e}dan, G.
\newblock Adversarial frontier stitching for remote neural network watermarking.
\newblock \emph{Neural Computing and Applications (NCA)}, 32\penalty0 (13):\penalty0 9233--9244, 2020.

\bibitem[Li et~al.(2023)Li, Wang, Ren, Sun, and Qiu]{li2023origin}
Li, L., Wang, P., Ren, K., Sun, T., and Qiu, X.
\newblock Origin tracing and detecting of llms.
\newblock \emph{arXiv preprint arXiv:2304.14072}, 2023.

\bibitem[Li et~al.(2022{\natexlab{a}})Li, Zhu, Jia, Bai, Jiang, Xia, and Cao]{both_li2022move}
Li, Y., Zhu, L., Jia, X., Bai, Y., Jiang, Y., Xia, S.-T., and Cao, X.
\newblock Move: Effective and harmless ownership verification via embedded external features.
\newblock \emph{arXiv preprint arXiv:2208.02820}, 2022{\natexlab{a}}.

\bibitem[Li et~al.(2022{\natexlab{b}})Li, Zhu, Jia, Jiang, Xia, and Cao]{white_li2022defending}
Li, Y., Zhu, L., Jia, X., Jiang, Y., Xia, S.-T., and Cao, X.
\newblock Defending against model stealing via verifying embedded external features.
\newblock In \emph{Proceedings of the AAAI Conference on Artificial Intelligence (AAAI)}, volume~36, pp.\  1464--1472, 2022{\natexlab{b}}.

\bibitem[Li(2023)]{BiLLA}
Li, Z.
\newblock Billa: A bilingual llama with enhanced reasoning ability.
\newblock \url{https://github.com/Neutralzz/BiLLa}, 2023.

\bibitem[Lin et~al.(2022)Lin, Lin, Liang, Liu, and Wang]{lin2023mpt}
Lin, K., Lin, C.-C., Liang, L., Liu, Z., and Wang, L.
\newblock Mpt: Mesh pre-training with transformers for human pose and mesh reconstruction.
\newblock \emph{arXiv preprint arXiv:2211.13357}, 2022.

\bibitem[Liu et~al.(2021)Liu, Weng, and Zhu]{white_liu2021residuals}
Liu, H., Weng, Z., and Zhu, Y.
\newblock Watermarking deep neural networks with greedy residuals.
\newblock In \emph{Proceedings of the International Conference on Machine Learning (ICML)}, pp.\  6978--6988. PMLR, 2021.

\bibitem[Lou et~al.(2022)Lou, Guo, Zhang, Zhang, and Liu]{gray_lou2021meets}
Lou, X., Guo, S., Zhang, T., Zhang, Y., and Liu, Y.
\newblock When nas meets watermarking: ownership verification of dnn models via cache side channels.
\newblock \emph{IEEE Transactions on Circuits and Systems for Video Technology (TCSVT)}, 2022.

\bibitem[Lukas et~al.(2019)Lukas, Zhang, and Kerschbaum]{lukas2019deep}
Lukas, N., Zhang, Y., and Kerschbaum, F.
\newblock Deep neural network fingerprinting by conferrable adversarial examples.
\newblock \emph{arXiv preprint arXiv:1912.00888}, 2019.

\bibitem[Mitchell et~al.(2023)Mitchell, Lee, Khazatsky, Manning, and Finn]{mitchell2023detectgpt}
Mitchell, E., Lee, Y., Khazatsky, A., Manning, C.~D., and Finn, C.
\newblock Detectgpt: Zero-shot machine-generated text detection using probability curvature.
\newblock In Krause, A., Brunskill, E., Cho, K., Engelhardt, B., Sabato, S., and Scarlett, J. (eds.), \emph{International Conference on Machine Learning, {ICML} 2023, 23-29 July 2023, Honolulu, Hawaii, {USA}}, volume 202 of \emph{Proceedings of Machine Learning Research}, pp.\  24950--24962. {PMLR}, 2023.
\newblock URL \url{https://proceedings.mlr.press/v202/mitchell23a.html}.

\bibitem[OpenAI(2022)]{openaichatgpt}
OpenAI.
\newblock Introducing chatgpt.
\newblock 2022.
\newblock URL \url{https://openai.com/blog/chatgpt}.

\bibitem[OpenAI(2023)]{ai-text-classifier}
OpenAI.
\newblock Ai classifier.
\newblock 2023.
\newblock URL \url{https://beta.openai.com/ai-text-classifier}.

\bibitem[Pan et~al.(2022)Pan, Yan, Zhang, and Yang]{fingerprinting_pan2022metav}
Pan, X., Yan, Y., Zhang, M., and Yang, M.
\newblock Metav: A meta-verifier approach to task-agnostic model fingerprinting.
\newblock In \emph{Proceedings of the 28th ACM SIGKDD Conference on Knowledge Discovery and Data Mining (SIGKDD)}, pp.\  1327--1336, 2022.

\bibitem[Penedo et~al.(2023)Penedo, Malartic, Hesslow, Cojocaru, Cappelli, Alobeidli, Pannier, Almazrouei, and Launay]{falcon}
Penedo, G., Malartic, Q., Hesslow, D., Cojocaru, R., Cappelli, A., Alobeidli, H., Pannier, B., Almazrouei, E., and Launay, J.
\newblock The {R}efined{W}eb dataset for {F}alcon {LLM}: outperforming curated corpora with web data, and web data only.
\newblock \emph{arXiv preprint arXiv:2306.01116}, 2023.
\newblock URL \url{https://arxiv.org/abs/2306.01116}.

\bibitem[Peng et~al.(2022)Peng, Li, Chen, Zhang, Zhu, and Xue]{fingerprinting_peng2022uapfp}
Peng, Z., Li, S., Chen, G., Zhang, C., Zhu, H., and Xue, M.
\newblock Fingerprinting deep neural networks globally via universal adversarial perturbations.
\newblock In \emph{Proceedings of the IEEE/CVF Conference on Computer Vision and Pattern Recognition (CVPR)}, pp.\  13430--13439, 2022.

\bibitem[Radford et~al.(2019)Radford, Wu, Child, Luan, Amodei, Sutskever, et~al.]{radford2019language}
Radford, A., Wu, J., Child, R., Luan, D., Amodei, D., Sutskever, I., et~al.
\newblock Language models are unsupervised multitask learners.
\newblock \emph{OpenAI blog}, 1\penalty0 (8):\penalty0 9, 2019.

\bibitem[Rouhani et~al.(2019)Rouhani, Chen, and Koushanfar]{white_rouhani2018deepsigns}
Rouhani, B.~D., Chen, H., and Koushanfar, F.
\newblock Deepsigns: an end-to-end watermarking framework for protecting the ownership of deep neural networks.
\newblock In \emph{ACM International Conference on Architectural Support for Programming Languages and Operating Systems (ASPLOS)}, 2019.

\bibitem[Sadasivan et~al.(2023)Sadasivan, Kumar, Balasubramanian, Wang, and Feizi]{sadasivan2023aigenerated}
Sadasivan, V.~S., Kumar, A., Balasubramanian, S., Wang, W., and Feizi, S.
\newblock Can ai-generated text be reliably detected?
\newblock \emph{arXiv preprint arXiv:2303.11156}, 2023.

\bibitem[Sakaguchi et~al.(2021)Sakaguchi, Bras, Bhagavatula, and Choi]{sakaguchi2021winogrande}
Sakaguchi, K., Bras, R.~L., Bhagavatula, C., and Choi, Y.
\newblock Winogrande: An adversarial winograd schema challenge at scale.
\newblock \emph{Communications of the ACM}, 64\penalty0 (9):\penalty0 99--106, 2021.

\bibitem[Sun et~al.(2024)Sun, Li, and Zhang]{sun2024zkllm}
Sun, H., Li, J., and Zhang, H.
\newblock zkllm: Zero knowledge proofs for large language models.
\newblock \emph{arXiv preprint arXiv:2404.16109}, 2024.

\bibitem[Taori et~al.(2023)Taori, Gulrajani, Zhang, Dubois, Li, Guestrin, Liang, and Hashimoto]{alpaca}
Taori, R., Gulrajani, I., Zhang, T., Dubois, Y., Li, X., Guestrin, C., Liang, P., and Hashimoto, T.~B.
\newblock Stanford alpaca: An instruction-following llama model.
\newblock \url{https://github.com/tatsu-lab/stanford_alpaca}, 2023.

\bibitem[Taylor et~al.(2022)Taylor, Kardas, Cucurull, Scialom, Hartshorn, Saravia, Poulton, Kerkez, and Stojnic]{taylor2022galactica}
Taylor, R., Kardas, M., Cucurull, G., Scialom, T., Hartshorn, A., Saravia, E., Poulton, A., Kerkez, V., and Stojnic, R.
\newblock Galactica: A large language model for science.
\newblock \emph{arXiv preprint arXiv:2211.09085}, 2022.

\bibitem[Team(2023)]{2023internlm}
Team, I.
\newblock Internlm: A multilingual language model with progressively enhanced capabilities.
\newblock \url{https://github.com/InternLM/InternLM}, 2023.

\bibitem[Tian(2023)]{Gptzero}
Tian, E.
\newblock Gptzero: An ai text detector.
\newblock 2023.
\newblock URL \url{https://gptzero.me/}.

\bibitem[Touvron et~al.(2023{\natexlab{a}})Touvron, Lavril, Izacard, Martinet, Lachaux, Lacroix, Rozi{\`e}re, Goyal, Hambro, Azhar, et~al.]{touvron2023llama}
Touvron, H., Lavril, T., Izacard, G., Martinet, X., Lachaux, M.-A., Lacroix, T., Rozi{\`e}re, B., Goyal, N., Hambro, E., Azhar, F., et~al.
\newblock Llama: Open and efficient foundation language models.
\newblock \emph{arXiv preprint arXiv:2302.13971}, 2023{\natexlab{a}}.

\bibitem[Touvron et~al.(2023{\natexlab{b}})Touvron, Martin, Stone, Albert, Almahairi, Babaei, Bashlykov, Batra, Bhargava, Bhosale, et~al.]{touvron2023llama2}
Touvron, H., Martin, L., Stone, K., Albert, P., Almahairi, A., Babaei, Y., Bashlykov, N., Batra, S., Bhargava, P., Bhosale, S., et~al.
\newblock Llama 2: Open foundation and fine-tuned chat models.
\newblock \emph{arXiv preprint arXiv:2307.09288}, 2023{\natexlab{b}}.

\bibitem[Uchida et~al.(2017)Uchida, Nagai, Sakazawa, and Satoh]{Uchida_2017}
Uchida, Y., Nagai, Y., Sakazawa, S., and Satoh, S.
\newblock Embedding watermarks into deep neural networks.
\newblock In \emph{Proceedings of the 2017 {ACM} on International Conference on Multimedia Retrieval}. {ACM}, jun 2017.
\newblock \doi{10.1145/3078971.3078974}.
\newblock URL \url{https://doi.org/10.1145%2F3078971.3078974}.

\bibitem[Wahby et~al.(2018)Wahby, Tzialla, Shelat, Thaler, and Walfish]{wahby2018doubly}
Wahby, R.~S., Tzialla, I., Shelat, A., Thaler, J., and Walfish, M.
\newblock Doubly-efficient zksnarks without trusted setup.
\newblock In \emph{2018 IEEE Symposium on Security and Privacy (SP)}, pp.\  926--943. IEEE, 2018.

\bibitem[Wang(2023)]{alpaca-lora}
Wang, E.~J.
\newblock \url{https://github.com/tloen/alpaca-lora}, 2023.

\bibitem[Wang et~al.(2020)Wang, Wu, Zhang, and Yao]{wang2020watermarking}
Wang, J., Wu, H., Zhang, X., and Yao, Y.
\newblock Watermarking in deep neural networks via error back-propagation.
\newblock \emph{Electronic Imaging}, 2020\penalty0 (4):\penalty0 22--1, 2020.

\bibitem[Wang \& Kerschbaum(2019)Wang and Kerschbaum]{wang2019attacks}
Wang, T. and Kerschbaum, F.
\newblock Attacks on digital watermarks for deep neural networks.
\newblock In \emph{ICASSP 2019-2019 IEEE International Conference on Acoustics, Speech and Signal Processing (ICASSP)}, pp.\  2622--2626. IEEE, 2019.

\bibitem[Wang \& Kerschbaum(2021)Wang and Kerschbaum]{white_wang2021riga}
Wang, T. and Kerschbaum, F.
\newblock Riga: Covert and robust white-box watermarking of deep neural networks.
\newblock In \emph{Proceedings of the Web Conference 2021 (WWW)}, pp.\  993--1004, 2021.

\bibitem[Workshop et~al.(2022)Workshop, Scao, Fan, Akiki, Pavlick, Ili{\'c}, Hesslow, Castagn{\'e}, Luccioni, Yvon, et~al.]{workshop2023bloom}
Workshop, B., Scao, T.~L., Fan, A., Akiki, C., Pavlick, E., Ili{\'c}, S., Hesslow, D., Castagn{\'e}, R., Luccioni, A.~S., Yvon, F., et~al.
\newblock Bloom: A 176b-parameter open-access multilingual language model.
\newblock \emph{arXiv preprint arXiv:2211.05100}, 2022.

\bibitem[Wu et~al.(2020)Wu, Liu, Yao, and Zhang]{none_wu2020watermarking}
Wu, H., Liu, G., Yao, Y., and Zhang, X.
\newblock Watermarking neural networks with watermarked images.
\newblock \emph{IEEE Transactions on Circuits and Systems for Video Technology (TCSVT)}, 31\penalty0 (7):\penalty0 2591--2601, 2020.

\bibitem[Wu et~al.(2023{\natexlab{a}})Wu, Pang, Shen, Cheng, and Chua]{wu2023llmdet}
Wu, K., Pang, L., Shen, H., Cheng, X., and Chua, T.-S.
\newblock Llmdet: A large language models detection tool.
\newblock \emph{arXiv preprint arXiv:2305.15004}, 2023{\natexlab{a}}.

\bibitem[Wu et~al.(2023{\natexlab{b}})Wu, Irsoy, Lu, Dabravolski, Dredze, Gehrmann, Kambadur, Rosenberg, and Mann]{wu2023bloomberggpt}
Wu, S., Irsoy, O., Lu, S., Dabravolski, V., Dredze, M., Gehrmann, S., Kambadur, P., Rosenberg, D., and Mann, G.
\newblock Bloomberggpt: A large language model for finance.
\newblock \emph{arXiv preprint arXiv:2303.17564}, 2023{\natexlab{b}}.

\bibitem[Xiang et~al.(2021)Xiang, Xie, Guo, Li, and Zhang]{xiang2021protecting}
Xiang, T., Xie, C., Guo, S., Li, J., and Zhang, T.
\newblock Protecting your nlg models with semantic and robust watermarks.
\newblock \emph{arXiv preprint arXiv:2112.05428}, 2021.

\bibitem[Xiong et~al.(2022)Xiong, Feng, Li, Zhang, and Qin]{fingerprinting_xiong2022neural}
Xiong, C., Feng, G., Li, X., Zhang, X., and Qin, C.
\newblock Neural network model protection with piracy identification and tampering localization capability.
\newblock In \emph{Proceedings of the 30th ACM International Conference on Multimedia (MM)}, pp.\  2881--2889, 2022.

\bibitem[Xu et~al.(2023{\natexlab{a}})Xu, Guo, Duan, and McAuley]{xu2023baize}
Xu, C., Guo, D., Duan, N., and McAuley, J.
\newblock Baize: An open-source chat model with parameter-efficient tuning on self-chat data.
\newblock \emph{arXiv preprint arXiv:2304.01196}, 2023{\natexlab{a}}.

\bibitem[Xu et~al.(2023{\natexlab{b}})Xu, Sun, Zheng, Geng, Zhao, Feng, Tao, and Jiang]{xu2023wizardlm}
Xu, C., Sun, Q., Zheng, K., Geng, X., Zhao, P., Feng, J., Tao, C., and Jiang, D.
\newblock Wizardlm: Empowering large language models to follow complex instructions.
\newblock \emph{arXiv preprint arXiv:2304.12244}, 2023{\natexlab{b}}.

\bibitem[Xu et~al.(2024)Xu, Wang, Ma, Koh, Xiao, and Chen]{xu2024instructional}
Xu, J., Wang, F., Ma, M.~D., Koh, P.~W., Xiao, C., and Chen, M.
\newblock Instructional fingerprinting of large language models.
\newblock \emph{arXiv preprint arXiv:2401.12255}, 2024.

\bibitem[Yadollahi et~al.(2021)Yadollahi, Shoeleh, Dadkhah, and Ghorbani]{yadollahi2021robust}
Yadollahi, M.~M., Shoeleh, F., Dadkhah, S., and Ghorbani, A.~A.
\newblock Robust black-box watermarking for deep neural network using inverse document frequency.
\newblock In \emph{2021 IEEE Intl Conf on Dependable, Autonomic and Secure Computing, Intl Conf on Pervasive Intelligence and Computing, Intl Conf on Cloud and Big Data Computing, Intl Conf on Cyber Science and Technology Congress (DASC/PiCom/CBDCom/CyberSciTech)}, pp.\  574--581. IEEE, 2021.

\bibitem[Yang et~al.(2022)Yang, Wang, and Wang]{fingerprinting_yang2022metafinger}
Yang, K., Wang, R., and Wang, L.
\newblock Metafinger: Fingerprinting the deep neural networks with meta-training.
\newblock In \emph{{Proceedings of the International Joint Conference on Artificial Intelligence (IJCAI)}}, 2022.

\bibitem[Zellers et~al.(2019)Zellers, Holtzman, Bisk, Farhadi, and Choi]{zellers2019hellaswag}
Zellers, R., Holtzman, A., Bisk, Y., Farhadi, A., and Choi, Y.
\newblock Hellaswag: Can a machine really finish your sentence?
\newblock In \emph{Proceedings of the 57th Annual Meeting of the Association for Computational Linguistics}, pp.\  4791--4800, 2019.

\bibitem[Zhang et~al.(2018)Zhang, Isola, Efros, Shechtman, and Wang]{zhang2018unreasonable}
Zhang, R., Isola, P., Efros, A.~A., Shechtman, E., and Wang, O.
\newblock The unreasonable effectiveness of deep features as a perceptual metric.
\newblock In \emph{Proceedings of the IEEE conference on computer vision and pattern recognition}, pp.\  586--595, 2018.

\bibitem[Zhang et~al.(2022)Zhang, Roller, Goyal, Artetxe, Chen, Chen, Dewan, Diab, Li, Lin, et~al.]{zhang2022opt}
Zhang, S., Roller, S., Goyal, N., Artetxe, M., Chen, M., Chen, S., Dewan, C., Diab, M., Li, X., Lin, X.~V., et~al.
\newblock Opt: Open pre-trained transformer language models.
\newblock \emph{arXiv preprint arXiv:2205.01068}, 2022.

\bibitem[Zhao et~al.(2020)Zhao, Hu, Liu, Ma, Chen, and Hassan]{fingerprinting_zhao2020afa}
Zhao, J., Hu, Q., Liu, G., Ma, X., Chen, F., and Hassan, M.~M.
\newblock Afa: Adversarial fingerprinting authentication for deep neural networks.
\newblock \emph{Computer Communications}, 150:\penalty0 488--497, 2020.

\bibitem[Zheng et~al.(2023{\natexlab{a}})Zheng, Chiang, Sheng, Zhuang, Wu, Zhuang, Lin, Li, Li, Xing, et~al.]{zheng2023judging}
Zheng, L., Chiang, W.-L., Sheng, Y., Zhuang, S., Wu, Z., Zhuang, Y., Lin, Z., Li, Z., Li, D., Xing, E., et~al.
\newblock Judging llm-as-a-judge with mt-bench and chatbot arena.
\newblock \emph{arXiv preprint arXiv:2306.05685}, 2023{\natexlab{a}}.

\bibitem[Zheng et~al.(2023{\natexlab{b}})Zheng, Xia, Zou, Dong, Wang, Xue, Wang, Shen, Wang, Li, Su, Yang, and Tang]{zheng2023codegeex}
Zheng, Q., Xia, X., Zou, X., Dong, Y., Wang, S., Xue, Y., Wang, Z., Shen, L., Wang, A., Li, Y., Su, T., Yang, Z., and Tang, J.
\newblock Codegeex: A pre-trained model for code generation with multilingual evaluations on humaneval-x.
\newblock In \emph{KDD}, 2023{\natexlab{b}}.

\bibitem[Zheng et~al.(2022)Zheng, Wang, and Chang]{fingerprinting_zheng2022nonrepudiable}
Zheng, Y., Wang, S., and Chang, C.-H.
\newblock A dnn fingerprint for non-repudiable model ownership identification and piracy detection.
\newblock \emph{IEEE Transactions on Information Forensics and Security}, 17:\penalty0 2977--2989, 2022.

\bibitem[Zhu et~al.(2023)Zhu, Chen, Shen, Li, and Elhoseiny]{zhu2023minigpt}
Zhu, D., Chen, J., Shen, X., Li, X., and Elhoseiny, M.
\newblock Minigpt-4: Enhancing vision-language understanding with advanced large language models.
\newblock \emph{arXiv preprint arXiv:2304.10592}, 2023.

\end{thebibliography}

\newpage
\appendix

\section{Additional related works}\label{Additional related works}
Deep neural network copyright protection methods can generally be divided into two categories: Watermarking and Fingerprinting.

\textbf{Watermarking}. There are three main types of watermarking methods. The first type involves embedding watermarks into model weights~\citep{chen2019deepmarks,white_wang2021riga,white_liu2021residuals,Uchida_2017}, hidden-layer activations~\citep{white_rouhani2018deepsigns}, gradients~\citep{white_li2022defending,both_li2022move}, model structures~\citep{gray_lou2021meets,lottery_chen2021you}, or extra components~\citep{passport_fan2019rethinking,passport_fan2021deepip}. These methods are typically white-box approaches and may potentially degrade the model's performance. The second type achieves watermarking by injecting triggers into the model to produce predefined outputs~\citep{black_adi2018turning,black_guo2018watermarking,black_le2020adversarial,black_chen2019blackmarks}; however, these methods often require fine-tuning or retraining the model. The third type relies on extractor subnetworks~\citep{none_wu2020watermarking,nlp_abdelnabi2021awl} or predefined rules~\citep{nlp_he2022dawnnlp,he2022cater} to embed watermarks into the model's output.

\textbf{Fingerprinting}. Fingerprinting methods are primarily divided into two categories. The first category involves copyright verification through the comparison of model weights~\citep{fingerprinting_jia2021proof} or their corresponding hash values~\citep{fingerprinting_zheng2022nonrepudiable,fingerprinting_chen2022perceptualhash,fingerprinting_xiong2022neural}, but these methods are limited to white-box scenarios and have only been tested on CNN-based visual models. The second category includes more recent works~\citep{fingerprinting_zhao2020afa,fingerprinting_pan2022metav,fingerprinting_yang2022metafinger,lukas2019deep,fingerprinting_peng2022uapfp} that construct DNN fingerprints by analyzing model behaviors on preset test cases. However, these methods often require additional models or data samples and may involve extra training.

\section{Details of Data Synthesis and Encoder Training }\label{Details of Data Synthesis and Encoder training}
\subsection{Data Synthesis}\label{Data Synthesis}
For the anchor data $\displaystyle \mM$:
Sample matrices $\displaystyle \mP_1, \displaystyle \mP_2, \displaystyle \mP_3$ from a standard normal distribution. Consider $\displaystyle \mP_1$ as $\hat{\displaystyle \mX}$, and $\displaystyle \mP_2, \displaystyle \mP_3$ as model parameter matrices, then
\begin{equation}\label{eq:anchor}
\displaystyle \mM=\displaystyle \mP_1\displaystyle \mP_2\displaystyle \mP_3\displaystyle \mP_1^T
\end{equation}
For positive data $\displaystyle \mM^+$:
Independently sample noises $\epsilon_i$ from a normal distribution $\mathcal{N}(0, \alpha)$, then
\begin{equation}\label{eq:positive}
\displaystyle \mP_i^+=\displaystyle \mP_i+\epsilon_i, \quad \displaystyle \mM^+=\displaystyle \mP_1^+\displaystyle \mP_2^+\displaystyle \mP_3^+\displaystyle \mP_1^{+^T}
\end{equation}
For negative data $\displaystyle \mM^-$:
Independently sample matrices $\displaystyle \mN_1, \displaystyle \mN_2, \displaystyle \mN_3$ from another standard normal distribution, then
\begin{equation}\label{eq:positive}
\displaystyle \mM^-=\displaystyle \mN_1\displaystyle \mN_2\displaystyle \mN_3\displaystyle \mN_1^T
\end{equation}
\subsection{Training the Encoder}

Note that we don't need to use any real LLM weights for training the encoder, as it only needs to learn a locality-preserving mapping between the input tensor and the output Gaussian vector. This ensures strict exclusivity between the training and test data. To construct the training data, we synthesize the matrix in each channel of $\displaystyle \mM$ on-the-fly, by randomly sampling 3 matrices $\displaystyle \mP_1,\displaystyle \mP_2,\displaystyle \mP_3$  and multiplying them together
as $\displaystyle \mP_1\displaystyle \mP_2\displaystyle \mP_3\displaystyle \mP_1^T$, as though they are model parameters. 

To learn locality-preserving mapping, we adopt contrastive learning. For a randomly sampled input $\displaystyle \mM$, its negative sample is given by another independently sampled tensor $\displaystyle \mM^-$. For its positive sample $\displaystyle \mM^+$, we perturb the content in each of $\displaystyle \mM$'s channel by adding small perturbation noises $\epsilon_i \in \mathcal{N}(0, \alpha)$ to the 3 matrices behind it. Here $\alpha$ is a hyperparameter determining the small variance. (c.f.~\appendixref{Data Synthesis} for detailed data synthesis process.) 

Subsequently, the contrastive loss $\mathcal{L}_C$ is given by:
\begin{equation}
\mathcal{L}_{C} =\left | (1- S_C(\displaystyle \mM,\displaystyle \mM^{+})) \right | +\left | S_C(\displaystyle \mM,\displaystyle \mM^{-}) \right |
\end{equation}
where $S_C(\cdot, \cdot)$ computes the cosine similarity between its two input matrices. 

To render the output vector to be Gaussian, we adopt the standard GAN~\citep{karras2019stylebased}  training scheme. We add a simple MLP as the discriminator $\displaystyle D$ that is trained to discriminate between real Gaussian vectors and the output vector $\displaystyle \vv$. In this setting, the encoder serves as the generator. During training, for every $m$ step, we alternate between training the discriminator and the generator. The discriminator loss $\mathcal{L}_D$ is thus given by
\begin{equation}
\mathcal{L}_{D} =\frac{1}{m}\sum_{i=1}^{m}\log_{}{\left (1-D\left ( \displaystyle \vv \right ) \right ) }
\end{equation}
While training the generator we also need to incorporate the contrastive learning loss. Thus the actual loss $\mathcal{L}$ for the training generator is a combination of $\mathcal{L}_C$ and $\mathcal{L}_D$.
\begin{equation}
\mathcal{L} = \displaystyle \mathcal{L}_{C} +\mathcal{L}_{D}
\end{equation}

\begin{table*}[t!]
  \centering
  \setlength{\tabcolsep}{5pt}
  \resizebox{1.0\columnwidth}{!}{
  \begin{tabular}{@{}lcccccccccc@{}}
  \toprule
  \textbf{Model} & \hspace{-0.2cm}BoolQ & \hspace{-0.2cm} HellaSwag \hspace{-0.2cm} & \hspace{-0.2cm}PIQA\hspace{-0.2cm} & \hspace{-0.3cm} WinoGrande \hspace{-0.2cm} & \hspace{-0.1cm}ARC-e & ARC-c&RACE&MMLU&Avg.\\
  \midrule
  \model   & 75.11        & 76.19 & 79.16 & 70.00 & 72.90       & 44.80&40.00&32.75&61.36  \\
 Alpaca  & 77.49        & 75.64 & 77.86 & 67.80 & 70.66       & 46.58 &43.16&41.13 &62.54     \\
   $+L_A(97.13)$& 45.44 & 31.16 & 67.63 & 48.70 & 49.03 & 34.13&22.78&23.13&40.25     \\
   $+L_A(87.21)$& 42.23 & 26.09 & 49.78 & 47.43 & 26.43 & 28.92&22.97&23.22&33.38     \\
  $+L_A(80.23)$& 39.05 & 26.40 & 49.95 & 48.30 & 26.52 & 28.75&22.97&23.98&33.24    \\
    $+L_A(77.56)$& 41.62 & 26.15 & 50.11 & 49.33 & 26.56 & 28.50&22.78&23.12&33.52 
    \\

  \bottomrule
  \end{tabular}
  }
  \caption{
  Detailed zero-shot performance on multiple standard benchmarks of the original \model, Alpaca, and the tuning model at different  $L_A$(PCS) values.
  }
  \label{tab:detailed performance}
\end{table*}
\section{Implementation Details}\label{Implementation Details}

\subsection{Training Settings}

In the training stage, we alternate training the discriminator and encoder every 10 steps. We set the batch size to 10, the initial learning rate to 0.0001, and introduce a noise intensity $\alpha$ of 0.16 for positive samples. After 8 epochs of training, we obtained the encoder used in our paper.

\subsection{Model Architecture}

For the encoder:
We used a convolution neural network (CNN) as the encoder.
The CNN encoder takes invariant terms $\displaystyle \mM \in \mathbb{R}^{4096 \times 4096 \times 6}$ as input and produces a feature vector $\displaystyle \vv$ as output. Our CNN encoder structure, as depicted in Figure \ref{fig: model_arch}, consists of the first four convolutional layers and the last mean pooling layer. The mean pooling layer simply calculates the average of the feature maps obtained from each channel, resulting in a feature vector $\displaystyle \vv$ with a length equal to the number of channels. The hyperparameters for the four convolutional layers are provided in the table below:

\begin{table}[!ht]
    \centering

    \begin{tabular}{c|c|c|c|c|c}
    \toprule
        CNN Layers & Input Channel & Output Channel&
        Kernel Size & Stride & Padding  \\
    \midrule 
        Layer 1 & 6 & 8 & 48 & 4 & 22   \\ 
        Layer 2 & 8 & 64 & 48 & 4 & 22  \\ 
        Layer 3 & 64 & 256 & 48 & 4 & 22  \\ 
        Layer 4 & 256 & 512 & 48 & 4 & 22   \\
    \bottomrule
    \end{tabular}
    
    \caption{
    Detailed hyperparameters of the stacked four convolutional layers.
    }
\end{table}

For the discriminator:
We utilize a simple 3-layer MLP as the discriminator. The 512-dimensional feature vector $\displaystyle \vv$ from the CNN encoder serves as fake data, while a 512-dimensional vector $\displaystyle \vx$ sampled from the standard normal distribution serves as real data. The discriminator processes $\displaystyle \vv$ and $\displaystyle \vx$, progressively reducing dimensionality through three linear layers, and finally outputs the probability of a sample being real after applying a sigmoid activation function. The sizes of the three linear layers are $\displaystyle \mW_1 \in \mathbb{R}^{512 \times 256}$, $\displaystyle \mW_2 \in \mathbb{R}^{256 \times 128}$, and $\displaystyle \mW_3 \in \mathbb{R}^{128 \times 1}$, respectively.

For the image generator: 
The pre-trained StyleGAN2 checkpoint we used can be found at:

\url{https://nvlabs-fi-cdn.nvidia.com/stylegan2-ada-pytorch/pretrained/afhqdog.pkl}

\section{Substitution Attack}\label{Substitution Attack}
\begin{figure*}[t]
\centering
\includegraphics[width=\textwidth]{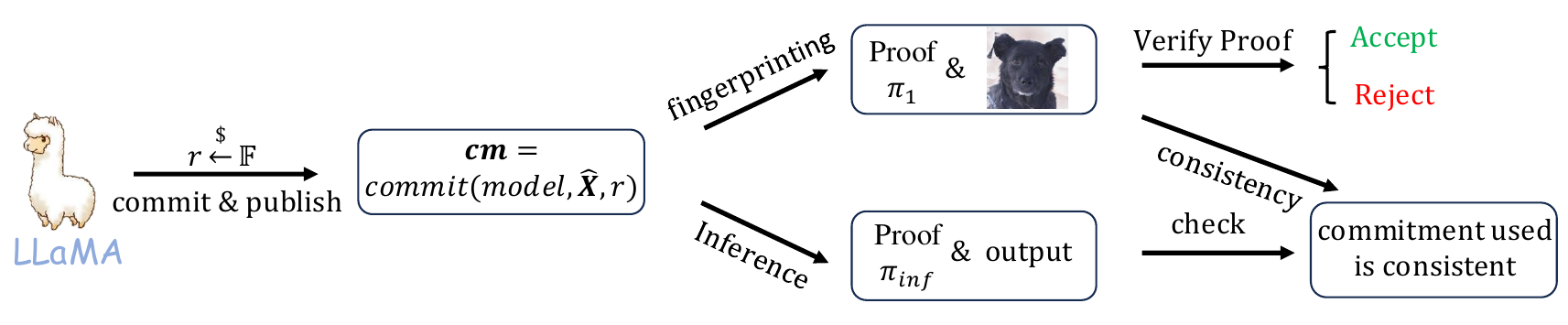} 
\caption{Flowchart for using commitment to defend against substitution attacks.}
\label{fig:subattack}
\end{figure*}
\vspace{-2mm}
\subsection{Training }
Substitution attack is a classic problem in cryptography. A conventional approach to address this issue is through cryptographic commitments~\citep{kate2010constant,wahby2018doubly}, which possess the dual properties of being binding and hiding:

\textbf{Binding}: This property ensures that it is computationally infeasible to find more than one valid opening for any given commitment, thereby preventing the substitution of the committed data.

\textbf{Hiding}: This ensures that the commitment itself discloses no information about the data it secures.

In our method, when a model developer wants to generate a fingerprint, they first commit to their model and publish this commitment. The binding property guarantees that no other model can match the same commitment, thereby preventing substitution attacks. All subsequent proof processes are carried out with this commitment, allowing anyone to verify if the model parameters used in calculations (such as fingerprinting or inferences) match those sealed within the commitment. (See~\autoref{fig:subattack} for an explanation of the process.)

For example, if a developer commits to model parameter A but uses a different model B for services, the public can request inference proofs for the model of the API for verification. Since the parameters used by model B inference are different from the parameters hidden in the commitment, the proof cannot pass the verification, substitution attacks will be revealed. For the Zero-Knowledge proof of LLM inference, we refer to~\cite{sun2024zkllm}, which provides an effective implementation. 
\section{StyleGAN2 Generator}\label{StyleGAN2 Generator}
StyleGAN2 is an improved model based on the style-based GAN architecture. One of its key enhancements is the incorporation of the perceptual path length (PPL) metric, which was originally introduced to quantify the smoothness of the mapping from the latent space to the output image. The PPL metric measures the average LPIPS distances~\citep{zhang2018unreasonable} between generated images under small perturbations in the latent space. Through the utilization of path length regularization, StyleGAN2 achieves enhanced reliability, consistency, and robustness, resulting in a smoother behavior of the generator. This regularization technique aligns with our objective of obtaining a locality-preserving generator.
\section{Open-sourced Independently Trained LLMs}\label{open-sourced LLMs}
In this experiment, we aimed to gather diverse models covering various parameter sizes. For the widely used LLaMA models, we included LLaMA-7B, 13B, 65B, LLaMA2-7B and 13B. We also incorporated models with similar architectures to LLaMA, such as InternLM-7B, Open\model-7B, and Baichuan-7B. To encompass a broader range of parameters, we expanded our collection to include GPT2-Large~\citep{radford2019language}, Cerebras-GPT-1.3B~\citep{dey2023cerebrasgpt}, Qwen-7B, 72B, Galactica-120B and even the largest Falcon-180B. Additionally, we considered models like MPT-7B, RedPajama-7B~\citep{together2023redpajama}, ChatGLM-6B~\citep{du2022glm}, Bloom-7.1B~\citep{workshop2023bloom}, ChatGLM2-6B, Pythia-6.9B and 12B~\citep{biderman2023pythia}, OPT-6.7B and 30B, and GPT-NeoX-20B~\citep{GPT-NeoX-20}, among other commonly used LLMs. Please refer to \autoref{tab:28 LLMs} for the ICSs between the 28 models.
\begin{sidewaystable}
\centering 
\scriptsize 
\setlength\tabcolsep{1.52pt} 
\begin{tabular}{l|cccccccccccccccccccccccccccc}
\toprule
\textbf{ICS} & GPT2 & CGPT & CLM & CLM2 & OPT6.7 & Py6.9 & MPT7 & Bai7 & Fal7 & Inte7 & OLM & LM7  & Qw7 & Bloom & LM27 & RedP & Py12 & LM213  & Bai13 & LM13 & Neox & LM30 & OPT30 & Fal40 & LM65 & Qw72 & Gal120 & Fal180 \\
\midrule
GPT2 & 100.00 & 18.06 & -0.67 & 0.01 & 5.50 & 0.03 & 0.53 & 0.16 & 0.30 & 0.21 & 0.05 & -0.15 & -0.07 & -0.45 & -0.04 & 0.03 & -0.04 & 0.09 & 0.30 & -0.09 & 0.11 & -0.09 & 3.36 & 0.79 & -0.24 & -0.09 & -0.37 & -1.35 \\
CGPT & 18.06 & 100.00 & -0.29 & 0.08 & 7.46 & 0.14 & 1.06 & 0.23 & 0.48 & 0.07 & 0.23 & -0.30 & 0.10 & -0.79 & -0.18 & 0.74 & 0.04 & 0.01 & 0.25 & 0.02 & 0.05 & -0.08 & 5.10 & 0.17 & 0.03 & -0.18 & -0.18 & -1.07 \\
CLM & -0.67 & -0.29 & 100.00 & 0.18 & -1.07 & -0.01 & -1.32 & -0.14 & -0.09 & -0.18 & -0.09 & 0.15 & 0.14 & 0.37 & -0.12 & 0.04 & 0.10 & 0.28 & -0.03 & -0.01 & -0.10 & -0.07 & -0.73 & 0.17 & 0.18 & 0.05 & -1.04 & 0.27 \\
CLM2 & 0.01 & 0.08 & 0.18 & 100.00 & -0.05 & 0.75 & -0.08 & 0.11 & 0.87 & 0.24 & 0.11 & 0.11 & 0.14 & 0.79 & 0.10 & 0.92 & 0.69 & 0.14 & 0.11 & 0.09 & 0.60 & -0.02 & -0.07 & 0.30 & -0.03 & 0.07 & -0.01 & -0.03 \\
OPT6.7 & 5.50 & 7.46 & -1.07 & -0.05 & 100.00 & 0.45 & 5.87 & 0.41 & 0.48 & -0.06 & -0.06 & -0.36 & -0.14 & -1.09 & 0.02 & 1.31 & 0.17 & -0.13 & 0.17 & -0.23 & 0.15 & -0.38 & 46.29 & 0.65 & -0.03 & -0.11 & -0.17 & -1.26 \\
Py6.9 & 0.03 & 0.14 & -0.01 & 0.75 & 0.45 & 100.00 & 0.13 & 0.01 & 0.66 & -0.06 & 0.04 & -0.00 & 0.01 & 0.55 & 0.02 & 2.37 & 1.58 & 0.02 & 0.01 & -0.02 & 1.41 & -0.00 & 0.29 & 0.23 & -0.01 & 0.02 & -0.01 & -0.04 \\
MPT7 & 0.53 & 1.06 & -1.32 & -0.08 & 5.87 & 0.13 & 100.00 & 0.32 & 0.44 & 0.13 & 0.10 & -0.13 & -0.12 & 0.83 & -0.10 & 0.62 & 0.40 & -0.18 & 0.52 & -0.03 & -0.28 & -0.49 & 1.10 & -1.23 & -0.33 & -0.12 & -0.61 & -0.82 \\
Bai7 & 0.16 & 0.23 & -0.14 & 0.11 & 0.41 & 0.01 & 0.32 & 100.00 & 0.13 & 0.21 & 0.21 & 0.32 & 0.41 & -0.13 & 0.35 & 0.09 & 0.00 & 0.22 & 0.42 & 0.28 & 0.04 & 0.10 & 0.21 & -0.08 & 0.10 & 0.31 & -0.16 & 0.01 \\
Fal7 & 0.30 & 0.48 & -0.09 & 0.87 & 0.48 & 0.66 & 0.44 & 0.13 & 100.00 & -0.06 & 0.04 & 0.08 & 0.13 & 0.48 & 0.23 & 0.84 & 0.62 & 0.05 & 0.16 & 0.01 & 0.54 & 0.19 & 0.39 & 1.68 & 0.05 & 0.19 & 0.01 & -11.07 \\
Inte7 & 0.21 & 0.07 & -0.18 & 0.24 & -0.06 & -0.06 & 0.13 & 0.21 & -0.06 & 100.00 & 0.18 & 0.03 & 0.48 & -0.01 & -0.13 & 0.02 & 0.02 & 0.36 & 0.13 & 0.08 & -0.00 & -0.31 & -0.64 & 0.08 & -0.29 & -0.26 & 0.00 & -0.01 \\
OLM & 0.05 & 0.23 & -0.09 & 0.11 & -0.06 & 0.04 & 0.10 & 0.21 & 0.04 & 0.18 & 100.00 & 0.32 & 0.32 & 0.09 & 0.39 & 0.06 & 0.03 & 0.23 & 0.35 & 0.27 & 0.05 & 0.19 & 0.01 & -0.04 & 0.06 & 0.32 & 0.08 & -0.06 \\
LM7 & -0.15 & -0.30 & 0.15 & 0.11 & -0.36 & -0.00 & -0.13 & 0.32 & 0.08 & 0.03 & 0.32 & 100.00 & 0.60 & 0.08 & 3.16 & 0.06 & 0.02 & 1.64 & 0.62 & 2.07 & 0.00 & 1.15 & 0.04 & -0.02 & 1.59 & 0.67 & 0.06 & 0.04 \\
Qw7 & -0.07 & 0.10 & 0.14 & 0.14 & -0.14 & 0.01 & -0.12 & 0.41 & 0.13 & 0.48 & 0.32 & 0.60 & 100.00 & 0.01 & 0.53 & -0.02 & 0.04 & 0.46 & 0.57 & 0.42 & -0.00 & -0.20 & -0.12 & -0.08 & 0.03 & 0.76 & 0.11 & -0.01 \\
Bloom & -0.45 & -0.79 & 0.37 & 0.79 & -1.09 & 0.55 & 0.83 & -0.13 & 0.48 & -0.01 & 0.09 & 0.08 & 0.01 & 100.00 & -0.09 & 0.35 & 0.48 & 0.11 & 0.03 & 0.07 & 0.41 & 0.02 & -0.68 & 0.05 & -0.08 & 0.01 & -0.00 & -0.18 \\
LM27 & -0.04 & -0.18 & -0.12 & 0.10 & 0.02 & 0.02 & -0.10 & 0.35 & 0.23 & -0.13 & 0.39 & 3.16 & 0.53 & -0.09 & 100.00 & -0.04 & -0.03 & 1.45 & 0.64 & 1.67 & 0.02 & 1.77 & 0.37 & -0.04 & 1.71 & 0.87 & 0.15 & 0.16 \\
RedP & 0.03 & 0.74 & 0.04 & 0.92 & 1.31 & 2.37 & 0.62 & 0.09 & 0.84 & 0.02 & 0.06 & 0.06 & -0.02 & 0.35 & -0.04 & 100.00 & 2.08 & -0.00 & -0.02 & 0.03 & 1.91 & -0.13 & 0.68 & 0.29 & 0.03 & 0.12 & 0.21 & -0.15 \\
Py12 & -0.04 & 0.04 & 0.10 & 0.69 & 0.17 & 1.58 & 0.40 & 0.00 & 0.62 & 0.02 & 0.03 & 0.02 & 0.04 & 0.48 & -0.03 & 2.08 & 100.00 & 0.04 & -0.01 & -0.02 & 1.27 & -0.02 & 0.08 & 0.30 & -0.04 & -0.03 & -0.03 & -0.00 \\
LM213 & 0.09 & 0.01 & 0.28 & 0.14 & -0.13 & 0.02 & -0.18 & 0.22 & 0.05 & 0.36 & 0.23 & 1.64 & 0.46 & 0.11 & 1.45 & -0.00 & 0.04 & 100.00 & 0.35 & 1.03 & -0.01 & -0.06 & -0.39 & -0.00 & 0.15 & 0.20 & -0.06 & 0.13 \\
Bai13 & 0.30 & 0.25 & -0.03 & 0.11 & 0.17 & 0.01 & 0.52 & 0.42 & 0.16 & 0.13 & 0.35 & 0.62 & 0.57 & 0.03 & 0.64 & -0.02 & -0.01 & 0.35 & 100.00 & 0.41 & -0.01 & 0.21 & 0.21 & -0.14 & 0.25 & 0.59 & 0.02 & -0.10 \\
LM13 & -0.09 & 0.02 & -0.01 & 0.09 & -0.23 & -0.02 & -0.03 & 0.28 & 0.01 & 0.08 & 0.27 & 2.07 & 0.42 & 0.07 & 1.67 & 0.03 & -0.02 & 1.03 & 0.41 & 100.00 & -0.01 & 0.39 & 0.13 & -0.12 & 0.88 & 0.37 & 0.07 & -0.04 \\
Neox & 0.11 & 0.05 & -0.10 & 0.60 & 0.15 & 1.41 & -0.28 & 0.04 & 0.54 & -0.00 & 0.05 & 0.00 & -0.00 & 0.41 & 0.02 & 1.91 & 1.27 & -0.01 & -0.01 & -0.01 & 100.00 & -0.00 & 0.14 & 0.34 & 0.02 & 0.03 & 0.11 & 0.01 \\
LM30 & -0.09 & -0.08 & -0.07 & -0.02 & -0.38 & -0.00 & -0.49 & 0.10 & 0.19 & -0.31 & 0.19 & 1.15 & -0.20 & 0.02 & 1.77 & -0.13 & -0.02 & -0.06 & 0.21 & 0.39 & -0.00 & 100.00 & 0.12 & 0.08 & 2.45 & 0.48 & -0.13 & 0.06 \\
OPT30 & 3.36 & 5.10 & -0.73 & -0.07 & 46.29 & 0.29 & 1.10 & 0.21 & 0.39 & -0.64 & 0.01 & 0.04 & -0.12 & -0.68 & 0.37 & 0.68 & 0.08 & -0.39 & 0.21 & 0.13 & 0.14 & 0.12 & 100.00 & 0.55 & 0.56 & 0.40 & -0.06 & -0.93 \\
Fal40 & 0.79 & 0.17 & 0.17 & 0.30 & 0.65 & 0.23 & -1.23 & -0.08 & 1.68 & 0.08 & -0.04 & -0.02 & -0.08 & 0.05 & -0.04 & 0.29 & 0.30 & -0.00 & -0.14 & -0.12 & 0.34 & 0.08 & 0.55 & 100.00 & -0.05 & -0.10 & 0.20 & 4.90 \\
LM65 & -0.24 & 0.03 & 0.18 & -0.03 & -0.03 & -0.01 & -0.33 & 0.10 & 0.05 & -0.29 & 0.06 & 1.59 & 0.03 & -0.08 & 1.71 & 0.03 & -0.04 & 0.15 & 0.25 & 0.88 & 0.02 & 2.45 & 0.56 & -0.05 & 100.00 & 0.44 & -0.13 & 0.02 \\
Qw72 & -0.09 & -0.18 & 0.05 & 0.07 & -0.11 & 0.02 & -0.12 & 0.31 & 0.19 & -0.26 & 0.32 & 0.67 & 0.76 & 0.01 & 0.87 & 0.12 & -0.03 & 0.20 & 0.59 & 0.37 & 0.03 & 0.48 & 0.40 & -0.10 & 0.44 & 100.00 & 0.07 & 0.09 \\
Gal120 & -0.37 & -0.18 & -1.04 & -0.01 & -0.17 & -0.01 & -0.61 & -0.16 & 0.01 & 0.00 & 0.08 & 0.06 & 0.11 & -0.00 & 0.15 & 0.21 & -0.03 & -0.06 & 0.02 & 0.07 & 0.11 & -0.13 & -0.06 & 0.20 & -0.13 & 0.07 & 100.00 & 0.19 \\
Fal180 & -1.35 & -1.07 & 0.27 & -0.03 & -1.26 & -0.04 & -0.82 & 0.01 & -11.07 & -0.01 & -0.06 & 0.04 & -0.01 & -0.18 & 0.16 & -0.15 & -0.00 & 0.13 & -0.10 & -0.04 & 0.01 & 0.06 & -0.93 & 4.90 & 0.02 & 0.09 & 0.19 & 100.00 \\
\bottomrule
\end{tabular}
\captionsetup{font=scriptsize} 
\caption{ICS between 28 open-sourced LLMs(774M to 180B): GPT2-Large (GPT2), Cerebras-GPT-1.3B (CGPT),  ChatGLM-6B (CLM), ChatGLM2-6B (CLM2), OPT-6.7B (OPT6.7), Pythia-6.9B (Py6.9), MPT-7B(MPT7), Baichuan-7B (Bai7), Falcon-7B (Fal7), InternLM-7B (Inte7), OpenLLaMA-7B (OLM), LLaMA-7B (LM7), Qwen-7B (Qw7), LLaMA2-7B2 (LM27), RedPajama-7B (RedP), Bloom-7B (Bloom), Pythia-12B (Py12), Baichuan-13B (Bai13), LLaMA-13B (LM13), GPT-NeoX-20B (Neox), LLaMA-30B (LM30), OPT-30B (OPT30), Falcon-40B (Fal40), LLaMA-65B (LM65), Qwen-72B (Qw72), Galactica-120B (Gala120), Falcon-180B (Fal180). Sorted left to right by parameter size from smallest to largest.}
\label{tab:28 LLMs}
\end{sidewaystable}

\section{Independently Trained LLMs in Smaller Scale}\label{Additional  Experiments}

To validate the uniqueness and stability of the parameter direction of LLMs trained from scratch, we independently trained GPT-NeoX-350M models on a subset of the Pile dataset~\citep{gao2020pile}. First, we examined whether different parameter initializations merely caused by global random seeds result in distinct parameter directions. Second, we explored the variation in the model's parameter vector direction during pretraining. 


\subsection{GPT-NeoX Models with Different Global Seeds}\label{sec:gptneox}
We investigated the impact of global random seeds on the model parameters' direction by independently training 4 GPT-NeoX-350M models on a subset of the Pile dataset. These models were trained using different global random number seeds while sharing the same architecture, training data batches, computational resources, and hyperparameters.

Subsequently, we computed the cosine similarities between these GPT-NeoX models' invariant terms, as shown in~\autoref{tab:seeds's confusion_matrix}. We generated fingerprints for these models, depicted in~\hyperref[4seeds]{Figure~\ref*{4seeds}}. The results revealed a noteworthy pattern: when GPT-NeoX models are trained from scratch, as long as the global random seed used for parameter initialization is different, it will lead to completely different parameter vector directions after pretraining.
Correspondingly, their fingerprints exhibited clear distinctions from each other.
\begin{table}[t!]
\begin{minipage}[b]{0.42\linewidth}
\centering
\setlength\tabcolsep{1pt}
\begin{tabular}{@{}c|cccc@{}}
\toprule
\textbf{ICS} & Seed=1 & Seed=2 & Seed=3 & Seed=4 \\ \midrule
Seed=1 & 100.00 & 2.08 & 2.23 & 2.08 \\
Seed=2 & 2.08 & 100.00 & 2.40 & 2.26 \\
Seed=3 & 2.23 & 2.40 & 100.00 & 2.29 \\
Seed=4 & 2.08 & 2.26 & 2.29 & 100.00 \\
\bottomrule
\end{tabular}
\captionof{table}{\text{ICS} values between GPT-NeoX models with different global seeds.}
\label{tab:seeds's confusion_matrix}
\end{minipage}
\hfill
\begin{minipage}[b]{0.56\linewidth}
\centering
\includegraphics[width=\textwidth,valign=c]{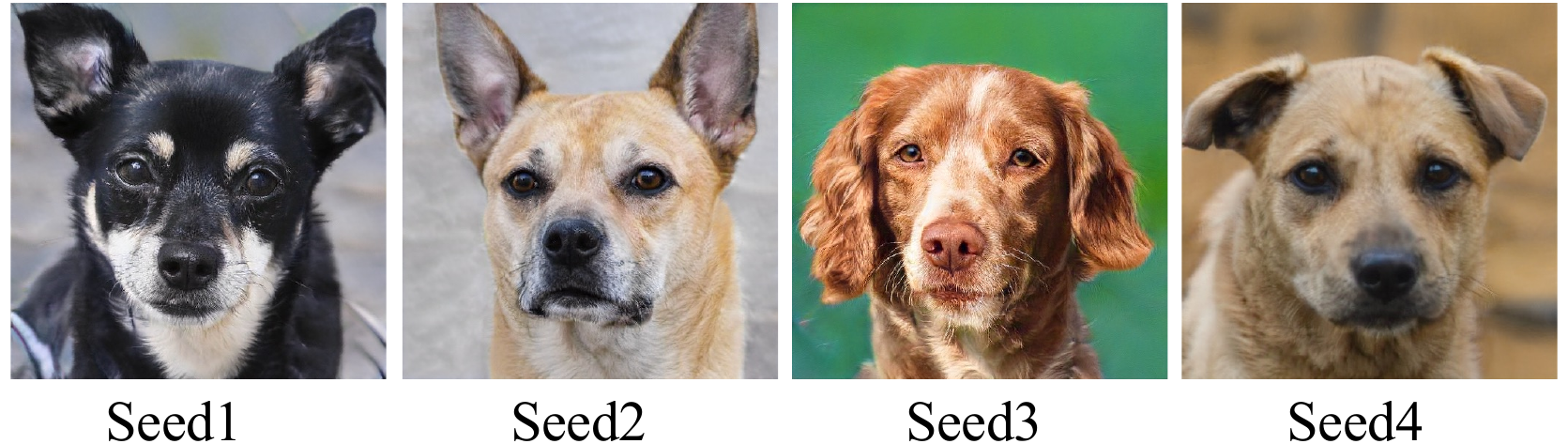}
\captionof{figure}{Fingerprints of GPT-NeoX models trained with varying global seeds.}
\label{4seeds}
\end{minipage}
\end{table}
\subsection{Model's Parameter Vector Direction's Variations During Pretraining}
In addition, we have explored the variation in the model's parameter vector direction during pretraining by comparing neighboring checkpoints of a model and calculating their cosine similarities. Specifically, we trained a GPT-NeoX-350M model on a subset of the Pile dataset for 360,000 steps and saved a checkpoint every 50k steps. As pretraining progresses, we observed a diminishing change in the model's parameter direction, leading to gradual stabilization, as shown in \autoref{tab:direction variation}. For larger models and more pretraining steps, we expect this phenomenon to be more pronounced, indicating that the parameter direction of LLMs tends to stabilize during pretraining.
\begin{table*}
  \centering
  \setlength{\tabcolsep}{5pt}
  \resizebox{1.0\columnwidth}{!}{
  \begin{tabular}{@{}lccccccc@{}}
  \toprule
  \textbf{Comparing CKPTs} & \hspace{-0.1cm}10k-60k & \hspace{-0.1cm} 60k-110k \hspace{-0.1cm} & \hspace{-0.1cm}110k-160k\hspace{-0.1cm} & \hspace{-0.1cm} 160k-210k \hspace{-0.1cm} & \hspace{-0.1cm}210k-260k\hspace{-0.1cm} & \hspace{-0.1cm}260k-310k\hspace{-0.1cm}&\hspace{-0.1cm}310k-360k\hspace{-0.1cm}\\
  \midrule
  Cosine similarity   & 56.23& 84.65 & 88.95 & 90.96 & 92.13       & 93.22&94.25  \\
  \bottomrule
  \end{tabular}
  }
  \caption{
  Cosine similarities between neighboring checkpoints (saved every 50k steps) of GPT-NeoX models during pretraining. 
  }
  \label{tab:direction variation}
\end{table*}
\begin{table*}
\small 
    \centering
    \begin{tabular}{@{}ccc@{}}
    \toprule
        \textbf{Offspring Model} & \textbf{Base Model} & \textbf{Detail}  \\
        \toprule
         Alpaca& \model-7B &  SFT on Stanford's instruction-following data \\
         Alpaca-Lora& \model-7B & SFT used the Lora training method  \\
         MiniGPT-4& \model-7B & multimodal model aligned on 5 million image-text pairs  \\
        Chinese-LLaMA& \model-7B & continued pretraining on Chinese corpus  \\
        Chinese-Alpaca& \model-7B & continued pretraining and finetuned on Chinese corpus \\
        Vicuna& \model-7B & SFT on around 125K user-shared conversations \\
        Baize& \model-7B & finetuned on 100k ChatGPT generated dialogs \\        
        Koala& \model-7B & SFT on dialogue data gathered from the web \\  
        WizardLM& \model-7B & trained on complex instructions data \\    
        MedAlpaca& \model-7B & finetuned on medical datasets \\             
        Beaver& \model-7B & underwent RLHF \\ 
        Guanaco& \model-7B & finetuned on nearly 600K multilingual
        dataset  \\     
        BiLLa& \model-7B & continued pretraining on a new language \\ 
        Falcon-40B-Instruct& Falcon-40B & finetuned on a mixture of Baize  \\ 
        Falcon-40B-SFT-Top1-560& Falcon-40B & SFT based on the OASST dataset~\citep{kopf2023openassistant} \\ 
        MPT-7B-Instruct& MPT-7B & finetuned on Dolly2 and HH-RLHF ~\citep{bai2022training} \\ 
        MPT-7B-StoryWriter& MPT-7B & finetuned with super long context length \\
        MPT-7B-Chat& MPT-7B & finetuned on a mixture of instruct datasets \\
        Qwen-7B-Chat& Qwen-7B & trained with alignment techniques \\
        Firefly-Qwen-7B& Qwen-7B & SFT by the Firefly project \\
        Baichuan-13B-Chat& Baichuan-13B & dialogue version \\
        Baichuan-13B-SFT& Baichuan-13B & bilingual instruction-tuned model \\
        InternLM-7B-Chat& InternLM-7B & optimized for dialogue use cases \\
        Firefly-InternLM& InternLM-7B & SFT by the Firefly project \\
        Qwen-72B-Chat& Qwen-72B & trained with alignment techniques \\
        OPT-IML-30B& OPT-30B & trained on 1500 tasks gathered from 8 NLP benchmarks  \\
        ChatGLM-fitness-RLHF& ChatGLM-6B & RLHF and SFT on millions data\\
        GPT-NeoXT-Chat& GPT-NeoX-20B & fine-tuned with 43 million high-quality instructions \\
        RedPajama-Chat& RedPajama-7B & fine-tuned on OASST1 and Dolly2~\citep{DatabricksBlog2023DollyV2} \\
        Bloomz-p3& Bloom-7B & finetuned on crosslingual task(P3) \\
        Bloomz-mt& Bloom-7B & multitask finetuned on xP3mt \\
        Falcon-180B-Chat& Falcon-180B & finetuned on a mixture of instruct datasets \\
        LLaMA2-7B-Chat& LLaMA2-7B & dialogue version \\
        LLaMA2-7B-32K& LLaMA2-7B & continued pretraining and SFT to enhance long-context capacity \\
        LLaMA2-function-calling& LLaMA2-7B & extends LLaMA2 model with function calling capabilities \\
        Llama2-Chinese-7B& LLaMA2-7B & aligned with Chinese dataset \\
        Vicuna2& LLaMA2-7B & fine-tuned on user-shared conversations \\
        LLaMA2-WikiChat& LLaMA2-7B & fine-tuned LLaMA-2 to retrieve data from Wikipedia \\
        Falcon-7B-Instruct& Falcon-7B & finetuned on a 250M tokens mixture of instruct/chat datasets \\
        Samantha-Falcon-7B& Falcon-7B & finetuned in philosophy, psychology, and personal relationships \\
        WizardLM-Falcon-7B& Falcon-7B & WizardLM trained on top of Falcon-7B \\        
        MPT-30B-Chat& MPT-30B & finetuned on a mixture of instruct datasets \\
        MPT-30B-Instruct& MPT-30B & finetuned on Dolly2 and HH-RLHF \\
        Baichuan-7B-SFT& Baichuan-7B & bilingual instruction-tuned model  \\
        Baichuan-7B-Chat& Baichuan-7B & dialogue version \\
        LLaMA2-13B-Chat& LLaMA2-13B & optimized for dialogue use cases \\
        LLaMA2-French& LLaMA2-13B & fine-tuned for answer questions in French \\
        LLaMA2-Estopia& LLaMA2-13B & focused on improving the dialogue and prose \\
        LLaMA2-Tiefighter& LLaMA2-13B & merging two different Lora's \\
        Llama2-Chinese-13B& LLaMA2-13B & aligned with Chinese dataset \\
        Nous-Hermes-Llama2-13B& LLaMA2-13B & fine-tuned on over 300,000 instructions \\
    \bottomrule
    \end{tabular}
    \caption{Detailed descriptions of all 51 offspring models.}
    \label{tab:all offspring model descriptions}
\end{table*}

\section{More Fingerprints}\label{more fingerprints}
\subsection{Offspring Models' Fingerprints}
\begin{figure*}[t]
\centering
\includegraphics[width=\textwidth]{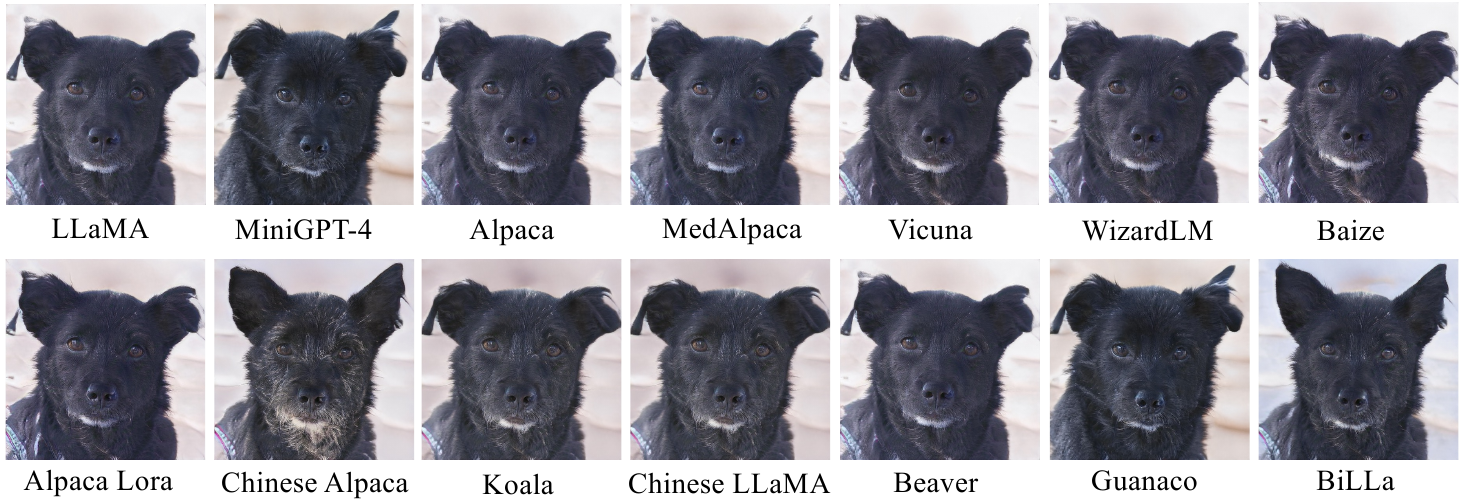} 
\caption{Fingerprints of LLaMA-7B and its offspring models.}
\label{exper1}
\end{figure*}

For LLaMA and its offspring models 
(\autoref{sec:fingerprints1})
, their fingerprints align with a similar fingerprint image of a Croatian sheepdog, exhibiting comparable poses, coat patterns, expressions, and backgrounds (\autoref{exper1}).

\begin{figure*}[t]
\centering
\includegraphics[width=\textwidth]{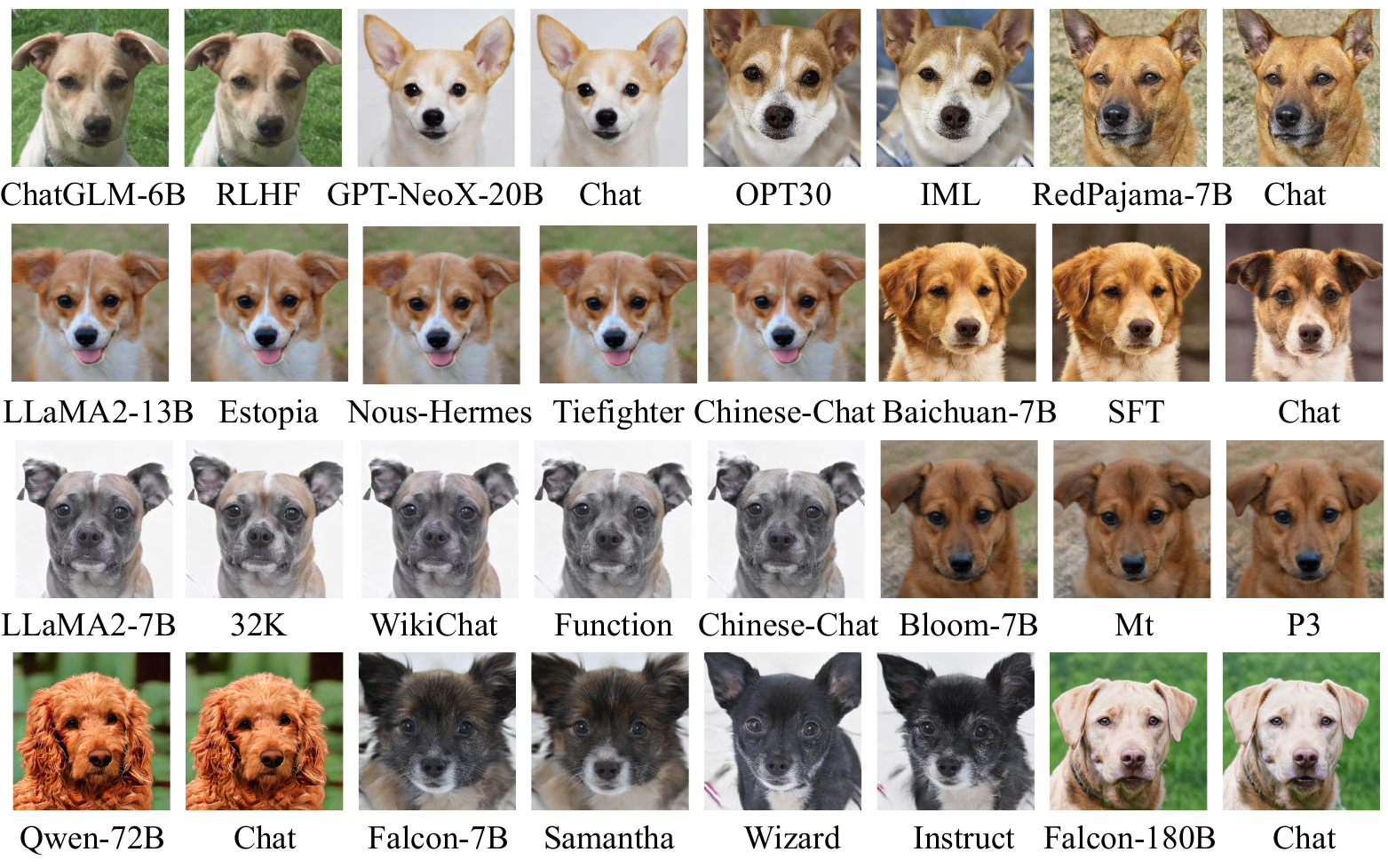} 
\caption{Fingerprints of the other offspring models and their base models.}
\label{Other offspring models' fingerprints}
\end{figure*}

In addition, the fingerprints of the rest offspring models listed in~\autoref{tab:all offspring model descriptions} and their respective base models are depicted in~\autoref{Other offspring models' fingerprints}. Offspring models' fingerprints still bear high similarity to their base models.

\subsubsection{28 Independently Trained LLMs' Fingerprints}

\begin{figure*}[t]
\centering
\includegraphics[width=\textwidth]{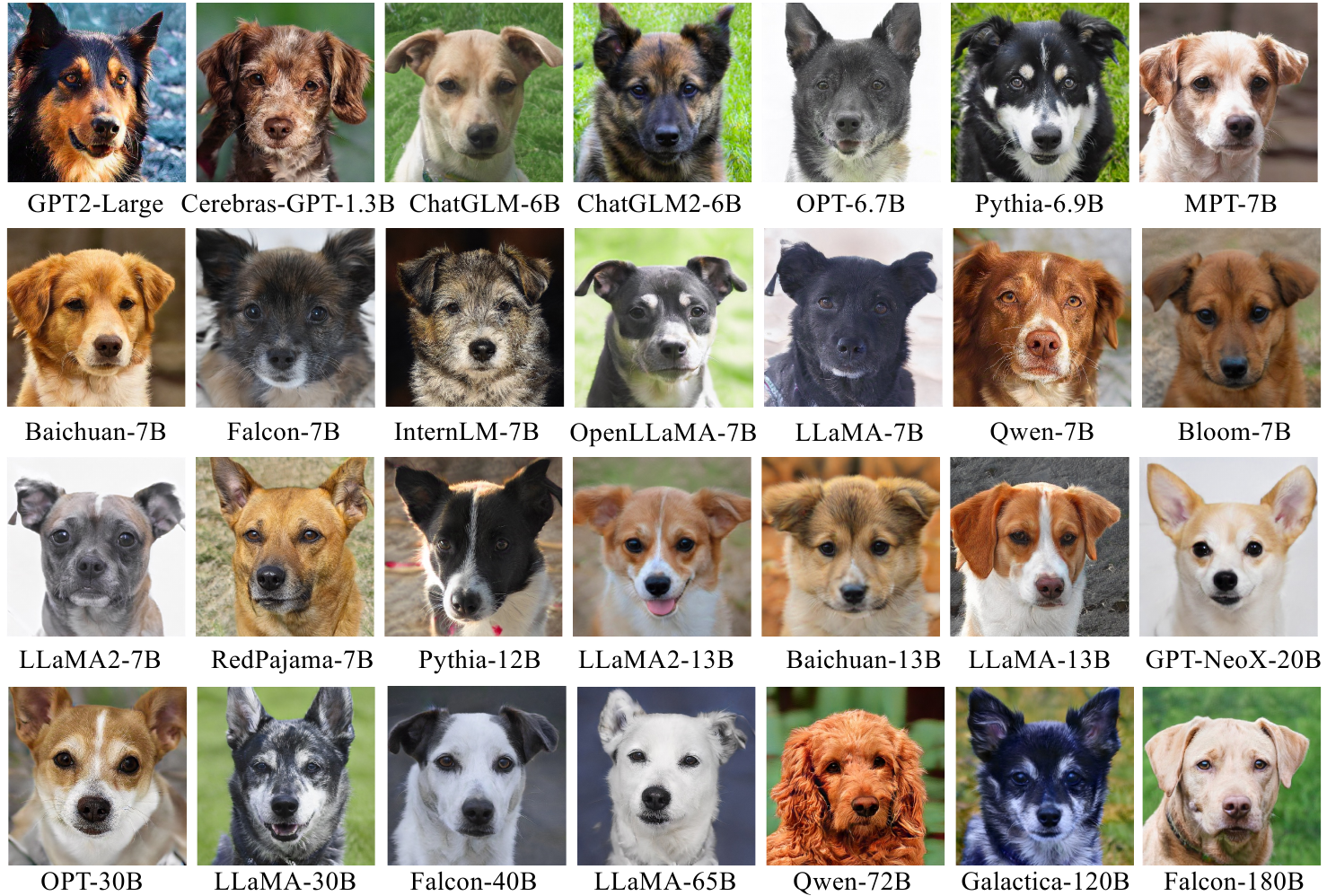} 
\caption{Fingerprints of 28 independently trained LLMs.}
\label{28 different base LLMs}
\end{figure*}

We also generate fingerprints for the 28 independently trained LLMs (\autoref{Diverse Set of Independently Trained LLMs}), as shown in~\autoref{28 different base LLMs}, their fingerprints exhibit high diversity, aligning with the distinct invariant terms of each model.



\clearpage

\section{Human Subject Study}\label{Human Subject Study}

\begin{figure*}[t!]
\centering
\includegraphics[width=\textwidth]{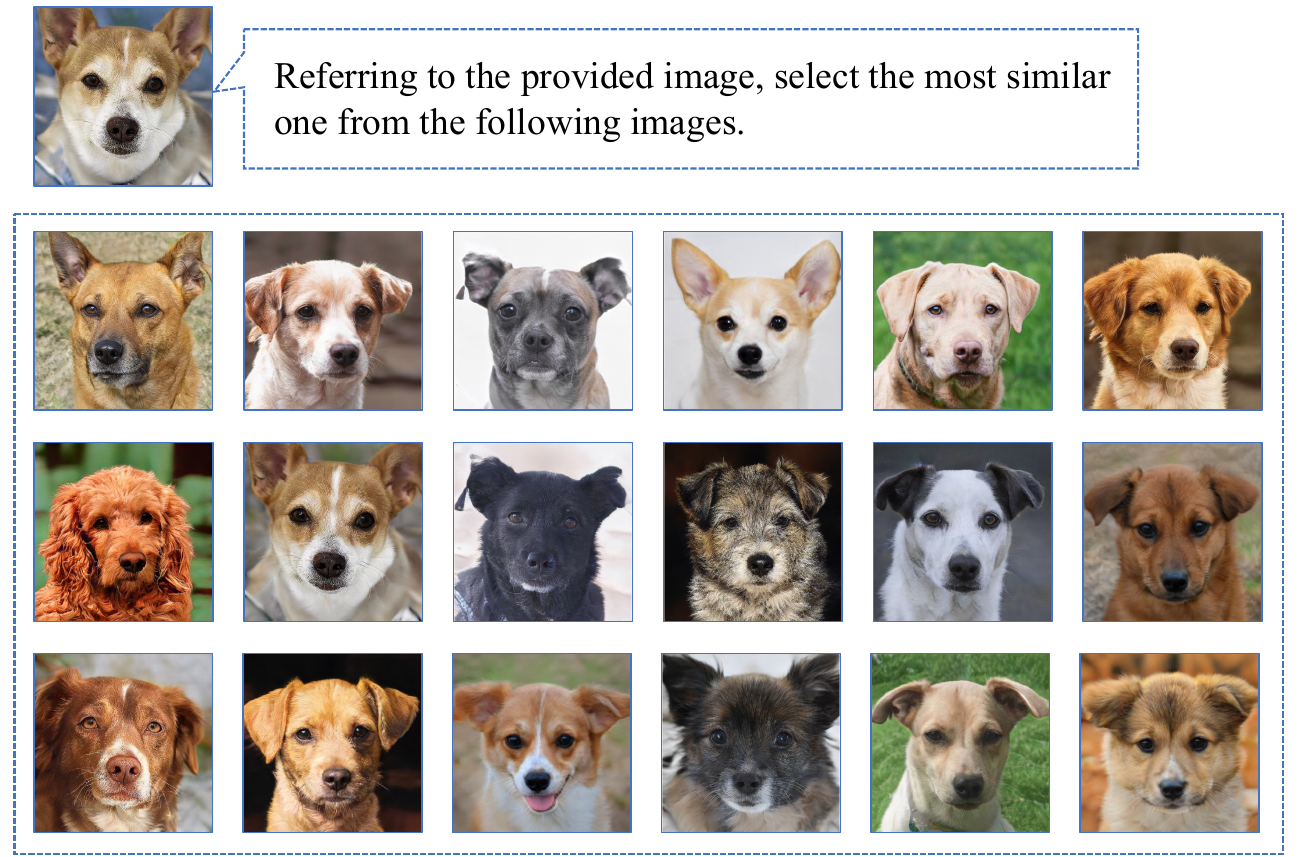} 
\caption{An illustration of a question in the human subject study. Participants were presented with the fingerprint of OPT-IML-30B (finetuned from OPT-30B) and asked to select the most similar image from the fingerprints of 18 distinct base models. Correct responses were counted only when participants precisely selected OPT-30B's fingerprint.}
\label{Example Question}
\end{figure*}

To evaluate the discrimination ability of our generated fingerprints, we generated fingerprints for the 51 offspring LLMs and their 18 base models (\autoref{tab:all offspring model descriptions}). We designed a single-choice test with 51 questions, each presenting an offspring model's fingerprint and asking participants to select the most similar image from the fingerprints of the 18 base models. 
 (c.f. ~\autoref{Example Question} for an example question and detailed description.) Conducted with \textbf{72} college-educated individuals, the test yielded a $\mathbf{94.74\%}$ \textbf{accuracy} rate, highlighting the discrimination ability and intuitive reflection of model similarity in our generated fingerprints.

\section{Experiments Compute Resources}\label{Experiments Compute Resources}
We trianed the CNN encoder for 2 hours using a single RTX4090. For extracting invariant terms and caculating cosine similarity, they only need a little cpu resources. The most compute resources are  consumed in reproduced baselines in~\autoref{Comparing to Latest Fingerprinting Methods}, in which we used 4 A100 40G for 8 days.


\section{Limitations}\label{Limitations}

Our method is only effective for transformer architecture LLMs, as the derivation of invariant terms is based on the transformer architecture. For non-transformer LLMs, our method may require modification to adapt to them.

\clearpage
 \section{More Illustrations of Baseline Comparison}\label{More Illustrations of Baseline Comparison}
Although Trap and IF are fingerprinting methods for LLMs, they differ significantly from our approach as they focus on protecting a specific LLM by eliciting predefined answers and then detecting them. In contrast, our work aims to safeguard base LLMs by identifying the underlying model of a given LLM.

Consequently, their evaluations are different from ours, and the results are not directly comparable. For example, Trap and IF report the proportion of prompts that successfully elicit predefined answers as the Fingerprint Success Rate (FSR). However, we cannot find a completely corresponding metric for comparison as we do not have predefined prompt-answer pairs. To provide a rough baseline comparison, we are compelled to report the accuracy of correctly identifying \model's offspring models as \model in~\autoref{Accuracy in Identify 51 Offspring Models' Base Model} as FSR.

\newpage
\clearpage

\end{document}